\documentclass[letterpaper, 10 pt, conference]{ieeeconf}
\IEEEoverridecommandlockouts
\usepackage{amsmath,amssymb,amsfonts}
\usepackage{algorithmic}
\usepackage{graphicx}
\usepackage{textcomp}
\usepackage{booktabs}
\usepackage{xcolor}
\def\BibTeX{{\rm B\kern-.05em{\sc i\kern-.025em b}\kern-.08em
    T\kern-.1667em\lower.7ex\hbox{E}\kern-.125emX}}
    
\usepackage{siunitx}
\usepackage{moresize}
\usepackage{tikz}
\usetikzlibrary{positioning}
\usetikzlibrary{calc}
\usetikzlibrary{arrows.meta}
\usetikzlibrary{shapes.geometric}
\usetikzlibrary{backgrounds}
\usetikzlibrary{spy}
\usepackage{blindtext}

\usepackage{subcaption}
\DeclareCaptionLabelSeparator{period}{.~~}
\DeclareCaptionLabelSeparator{colon}{:~~}
\DeclareCaptionLabelFormat{parens}{(#2)}
\makeatletter
\let\NAT@parse\undefined
\makeatother
\usepackage[bookmarks=false]{hyperref}
\usepackage{multirow}
\usepackage{nicefrac}
\usepackage{placeins}

\usepackage[noadjust]{cite}

\usepackage{lipsum}

\usepackage{dblfloatfix}

\usepackage{fancyhdr}

\newcommand{\uproman}[1]{\uppercase\expandafter{\romannumeral#1}}

\newcommand{\sIRGB}{\mathbf{I}_{\text{RGB}}}                    %
\newcommand{\sMi}{\mathbf{M}_{i}}                               %
\newcommand{\setM}{\mathbb{M}}                                  %

\newcommand{\seI}{\mathbf{e}^I}
\newcommand{\seMi}{\mathbf{e}^{\mathbf{M}_i}}
\newcommand{\steMi}{\tilde{\mathbf{e}}^{\mathbf{M}_i}}
\newcommand{\sY}{\mathbf{Y}}
\newcommand{\sYh}{\hat{\mathbf{Y}}}
\newcommand{\sYxy}{\mathbf{y}_{y,x}}
\newcommand{\sYhxy}{\hat{\mathbf{y}}_{y,x}}

\begin{document}

\title{\LARGE \bf
Efficient Prediction of Dense Visual Embeddings via Distillation and RGB-D Transformers
}
\def\theauthors{Söhnke Benedikt Fischedick, Daniel Seichter, Benedict Stephan, Robin Schmidt, and Horst-Michael Gross}
\author{\theauthors%
\thanks{Authors are with Neuroinformatics and Cognitive Robotics Lab, Technische Universit\"at Ilmenau, 98693 Ilmenau, Germany. Contact: \hfill\newline
{\scriptsize{\tt soehnke.fischedick@tu-ilmenau.de}, ORCID: {\tt
\href{https://orcid.org/0000-0001-8447-0584}{0000-0001-8447-0584}}}}%
\thanks{This work has received funding from Carl-Zeiss-Stiftung to the project \textit{Co-Presence of Humans and Interactive Companions for Seniors} (CO-HUMANICS) running 6/21-5/26. \href{http://www.co-humanics.de}{http://www.co-humanics.de}.}%
}

\maketitle
\newboolean{isarxiv}
\setboolean{isarxiv}{true}
\ifthenelse{\boolean{isarxiv}}{%
    \renewcommand{\headrulewidth}{0pt}
    \fancypagestyle{fancyfirstpage}{%
        \fancyhf{}%
        \addtolength{\topmargin}{-20pt}%
        \addtolength{\headheight}{20pt}%
        \fancyhead[C]{%
            \footnotesize%
            \textcolor{gray}{%
                © 2025 IEEE.
                Personal use of this material is permitted.
                Permission from IEEE must be obtained for all other uses, in any current or future media, including reprinting/republishing this material for advertising or promotional purposes, creating new collective works, for resale or redistribution to servers or lists, or reuse of any copyrighted component of this work in other works.
                DOI:\href{https://doi.org/10.1109/IROS60139.2025.11245809}{10.1109/IROS60139.2025.11245809}
            }%
        }%
        \fancyfoot[C]{%
            \footnotesize%
            \textcolor{gray}{\thepage}%
        }%
    }%
    \fancypagestyle{fancypage}{%
        \fancyhf{}%
        \fancyfoot[C]{%
            \footnotesize%
            \textcolor{gray}{\thepage}%
        }%
    }%
    \thispagestyle{fancyfirstpage}%
    \pagestyle{fancypage}%
}{%
    \thispagestyle{empty}%
    \pagestyle{empty}%
}%

\begin{abstract}
In domestic environments, robots require a comprehensive understanding of their surroundings to interact effectively and intuitively with untrained humans.
In this paper, we propose DVEFormer~--~an efficient RGB-D Transformer-based approach that predicts dense text-aligned visual embeddings~(DVE) via knowledge distillation.
Instead of directly performing classical semantic segmentation with fixed predefined classes, our method uses teacher embeddings from Alpha-CLIP to guide our efficient student model DVEFormer in learning fine-grained pixel-wise embeddings. 
While this approach still enables classical semantic segmentation, e.g., via linear probing, it further enables flexible text-based querying and other applications, such as creating comprehensive 3D maps.
Evaluations on common indoor datasets demonstrate that our approach achieves competitive performance while meeting real-time requirements, operating at 26.3$\,$FPS for the full model and 77.0$\,$FPS for a smaller variant on an NVIDIA Jetson AGX Orin.
Additionally, we show qualitative results that highlight the effectiveness and possible use cases in real-world applications.
Overall, our method serves as a drop-in replacement for traditional segmentation approaches while enabling flexible natural-language querying and seamless integration into 3D mapping pipelines for mobile robotics.
\end{abstract}

\section{Introduction}
\label{sec:introduction}
Mobile robots that operate autonomously in complex indoor environments require a detailed understanding of the environment in which they operate.
This scene knowledge should be comprehensive and universal, i.e., it should capture details of relevant objects in the scene beyond a fixed predefined class spectrum (often referred to as closed set), to be applicable in various environments.

In our ongoing research project CO-HUMANICS~\cite{Cohumanics-Fischedick-ISR-2023, cohuamnics-roman-2024, cohumanics-Conde2024}, a mobile robot with telepresence features is operating in an older adults' apartment and can be accessed and controlled remotely by family and friends to bridge physical distance between people.
In this scenario, the operator is often untrained, and, therefore, controlling the robot must be as intuitive as possible.
For example, when specifying a target position for the robot, it would be favorable to do this with natural-language identifiers beyond a fixed spectrum of predefined classes.
Fig.~\ref{fig:eye-catcher} visualizes this aspect for the example command \textit{``drive to the armchair``}.
This command requires the robot to not only identify an armchair as a distinct object, but also to differentiate it from other types of chairs.

\newcommand{\imagesize}{2cm}%
\newcommand{\ndtimagesize}{2.5cm}%
\newcommand{\inputimagesize}{2cm}%
\newcommand{\subinputimagesize}{0.8cm}%
\definecolor{emsaformer_encoder_color}{HTML}{d62727}%
\definecolor{emsaformer_context_color}{HTML}{ffc266}%
\definecolor{emsaformer_decoder_color}{HTML}{6cd56a}%
\definecolor{query_color}{HTML}{FFD6B8}%
\definecolor{framebased_color}{HTML}{E7E7E7}%
\definecolor{dveformer_box_color}{HTML}{6bff5f}%
\definecolor{application_arrow_color}{HTML}{fff1e7}%
\definecolor{data_arrow_color}{HTML}{d9d9d9}%
\definecolor{chair_color}{HTML}{800080}%
\definecolor{floor_color}{HTML}{008000}%
\def\volume[#1](#2)(#3, #4, #5, #6){
    \node[#1, draw, minimum width=#4, minimum height=#3] (#2) {};
    \draw[dashed] (#2.south west) -- ++(#5, #6);
    \draw (#2.south east) -- ++(#5, #6);
    \draw (#2.north west) -- ++(#5, #6);
    \draw (#2.north east) -- ++(#5, #6);
    \draw (#2.south east)++(#5, #6) -- ++(0, #3);
    \draw[dashed] (#2.south west)++(#5, #6) -- ++(0, #3);
    \draw[dashed] (#2.south west)++(#5, #6) -- ++(#4, 0);
    \draw (#2.north west)++(#5, #6) -- ++(#4, 0);
}%
\begin{figure}%
    \vspace{0.5mm}%
    \centering%
    \tikzset{image/.style={inner sep=0}}%
    \tikzset{ndtimage/.style={inner sep=0}}%
    \tikzset{emsaformer/.style={minimum width=2cm, minimum height=1cm, fill=red, font=\scriptsize, draw, trapeziod}}%
    \tikzset{prompt/.style={font=\scriptsize, gray}}%
    \tikzset{emsaformer_encoder/.style={fill=emsaformer_encoder_color, minimum width=1.8cm, minimum height=1.5cm, trapezium, trapezium angle=70, rotate=-90, draw, font=\scriptsize}}%
    \tikzset{emsaformer_decoder/.style={fill=emsaformer_decoder_color, minimum width=1.8cm, minimum height=0.5cm, trapezium, trapezium angle=75, rotate=90, draw}}%
    \tikzset{emsaformer_context/.style={fill=emsaformer_context_color, font=\scriptsize, draw, inner sep=3px}}%
    \tikzset{dveformer_box/.style={dveformer_box_color, line width=0px, opacity=0.15}}%
    \begin{tikzpicture}%
        \tikzset{trapezium stretches=true}%
        \coordinate (top_right) at (0.98\hsize, 6);%

        \node[anchor=south west, image] (ndt_example_1) at (0, 0) {%
            \begin{tikzpicture}%
                \node[rotate=0] {%
                    \includegraphics[width=\ndtimagesize]{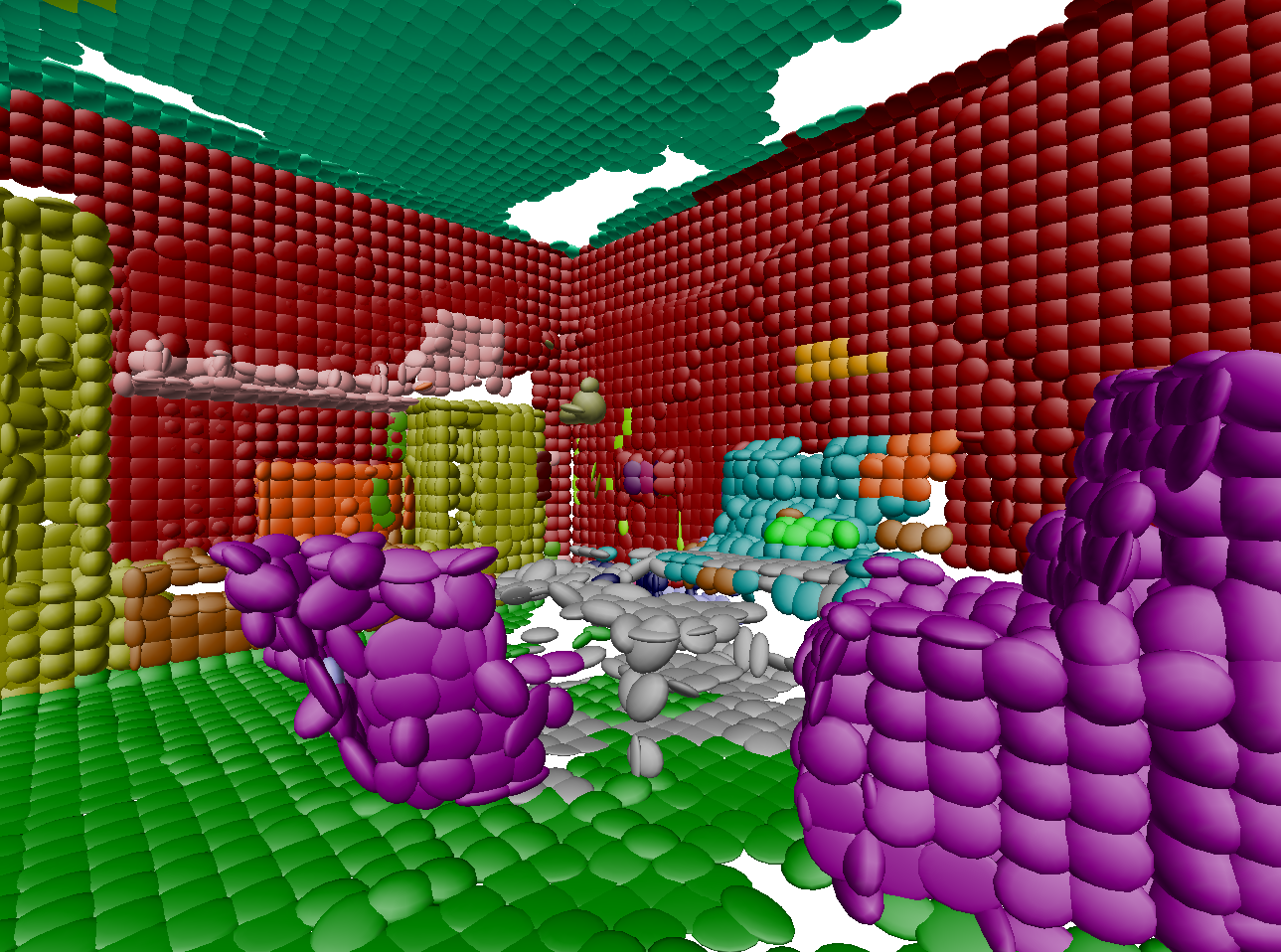}%
                };%
            \end{tikzpicture}%
        };%
        \node[ndtimage, right=0.4 of ndt_example_1] (ndt_example_2) {%
            \begin{tikzpicture}%
                \node[rotate=0] {%
                    \includegraphics[width=\ndtimagesize]{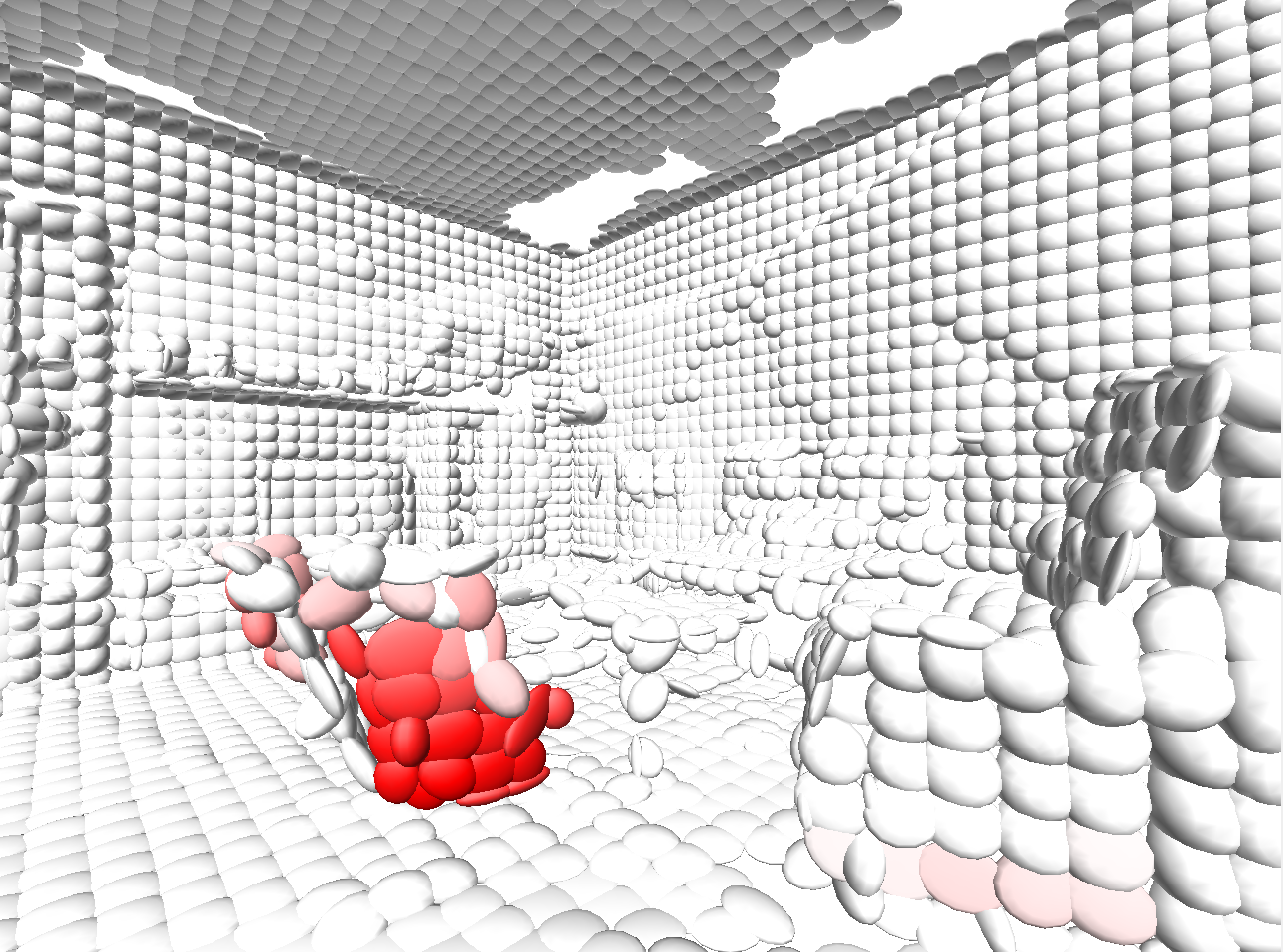}%
                };%
            \end{tikzpicture}%
        };%
        \node[ndtimage, right=0.1 of ndt_example_2] (ndt_example_3) {%
            \begin{tikzpicture}%
                \node[rotate=0] {%
                    \includegraphics[width=\ndtimagesize]{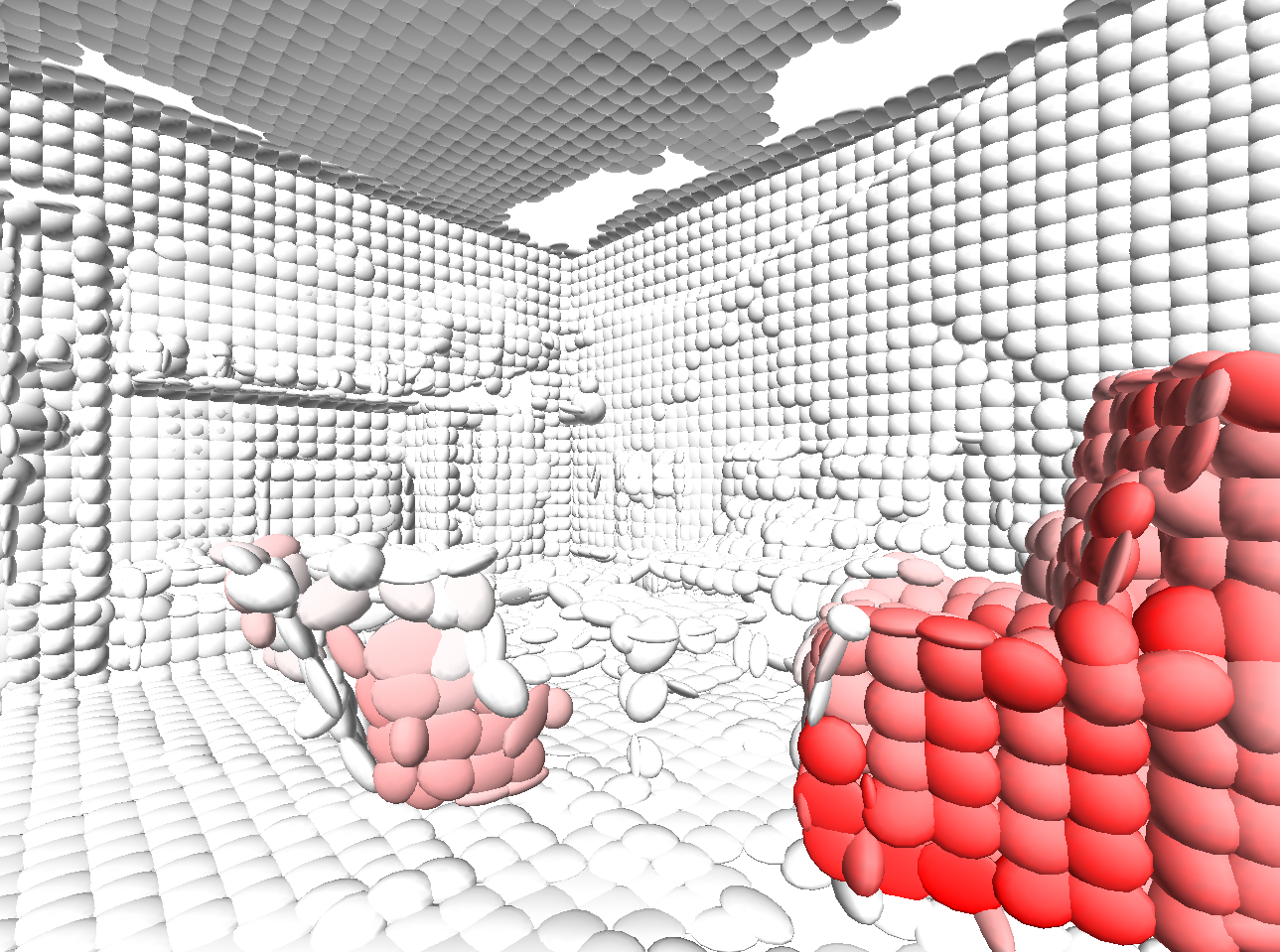}%
                };%
            \end{tikzpicture}%
        };%

        \node[above=0.1 of ndt_example_1, image] (img_example_1) {%
            \includegraphics[width=\imagesize]{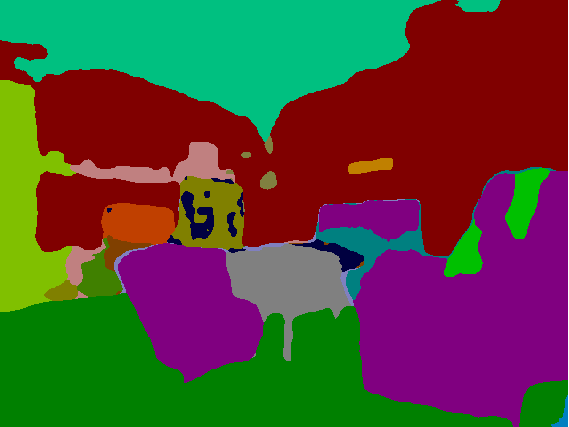}%
        };%
        \node[above=0.1 of ndt_example_2, image, shift={(0.1, 0)}] (img_example_2) {%
            \includegraphics[width=\imagesize]{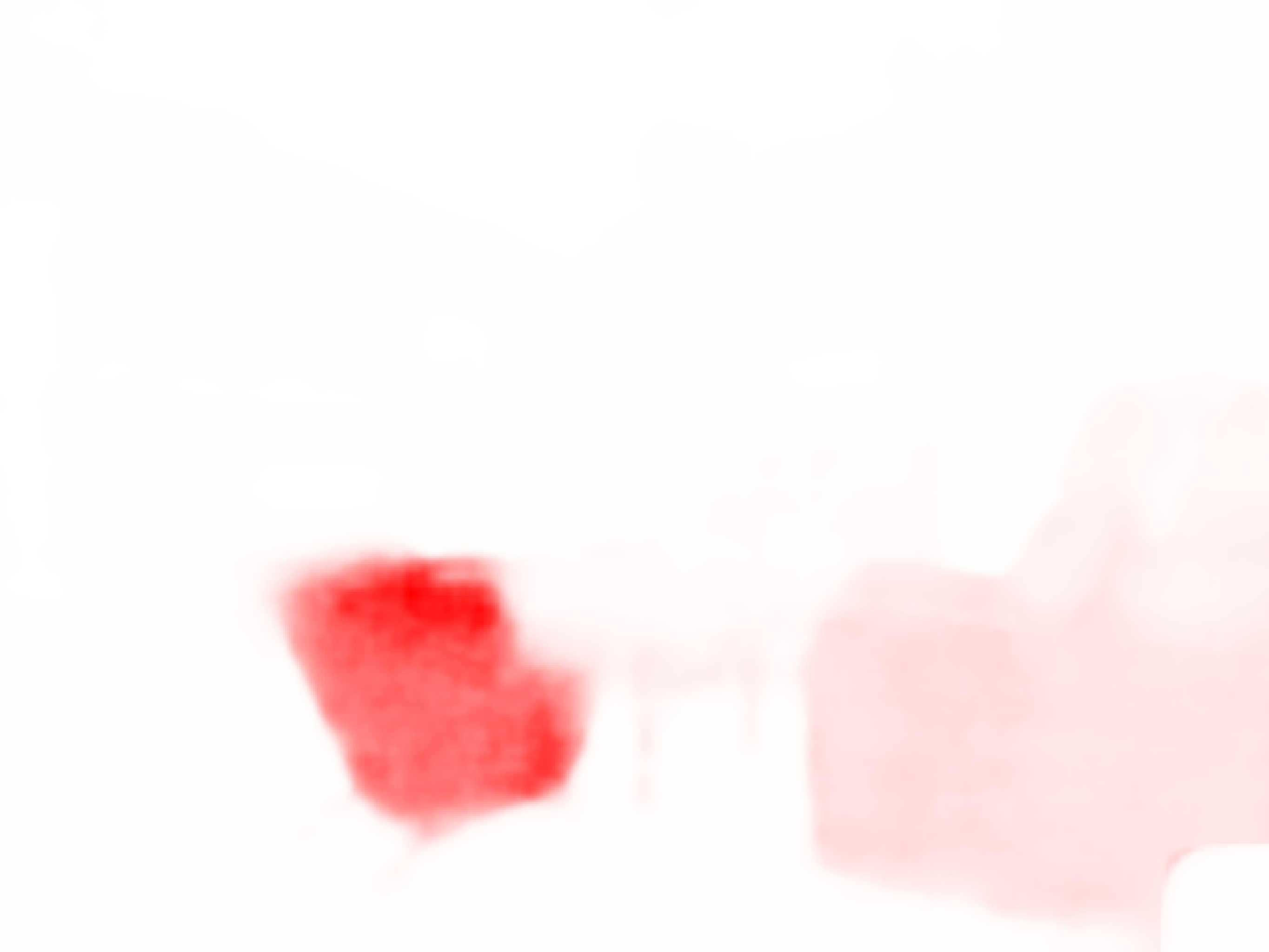}%
        };%
        \node[above=0.1 of ndt_example_3.north, anchor=south, image, shift={(0.1, 0)}] (img_example_3){%
            \includegraphics[width=\imagesize]{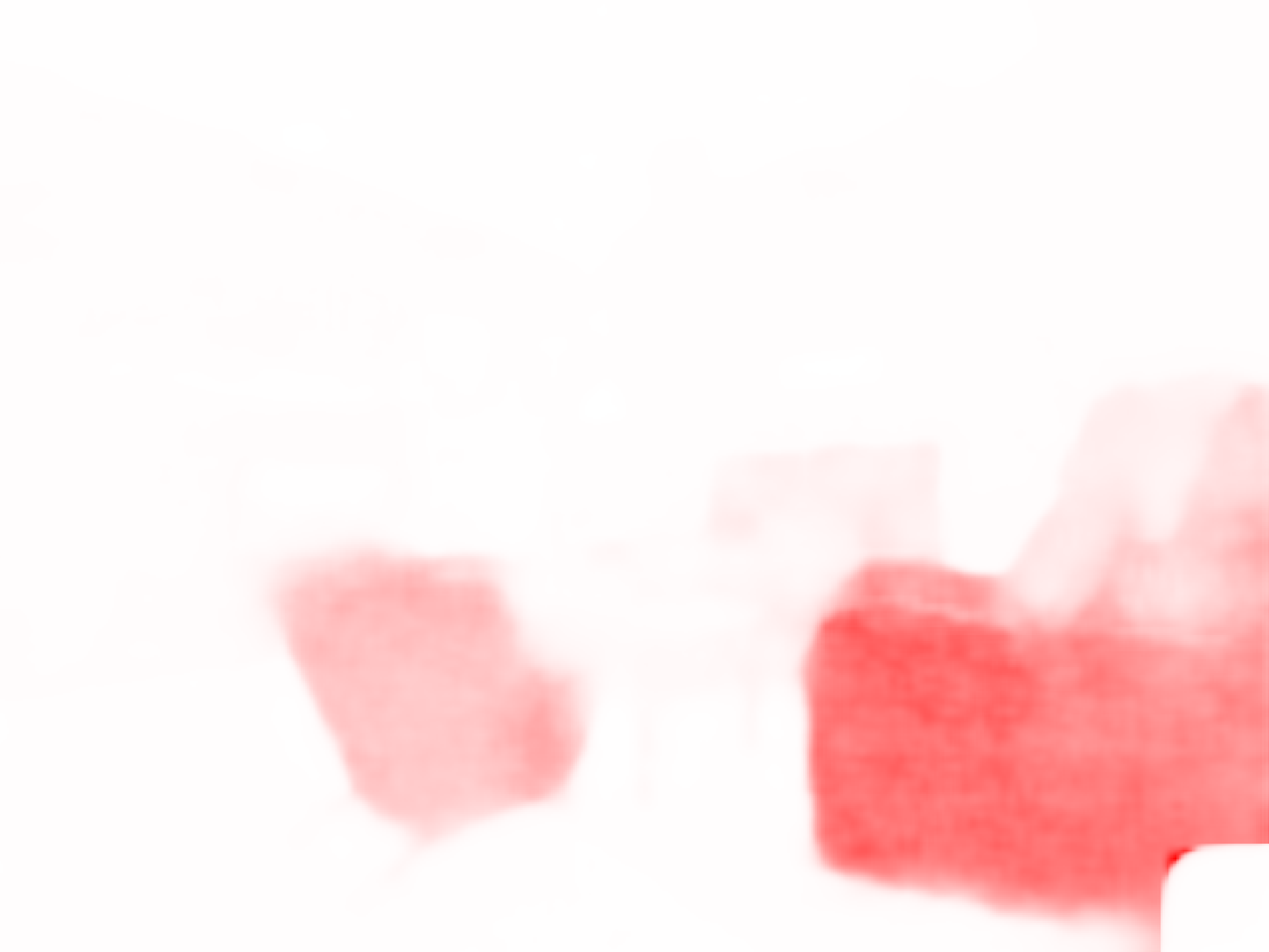}%
        };%

        \node[left=0 of img_example_2, prompt] {\rotatebox{90}{``Round chair``}};%
        \node[left=0 of img_example_3, prompt] {\rotatebox{90}{``Armchair``}};%

        \draw let%
            \p1=(top_right),%
            \p2=($(img_example_1)-(0.4, 0)$)%
        in%
            node[image, anchor=north] (rgb2) at (\x2, \y1) {\includegraphics[width=\inputimagesize]{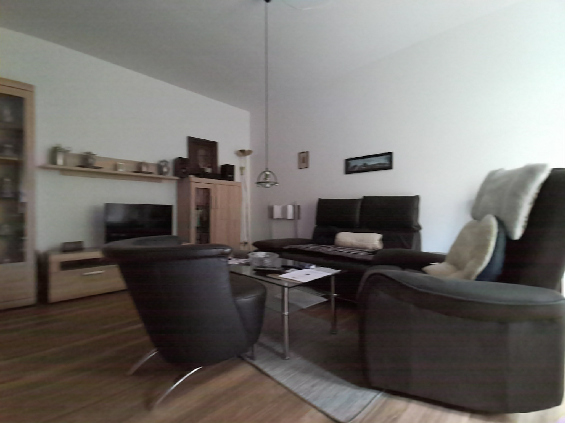}};%
        \node[font=\scriptsize, gray, left= 0.05 of rgb2.west, rotate=90, anchor=south, inner sep=0] {%
            RGB-D Input%
        };%

        \begin{scope}[on background layer]%
            \node[above right=0.18 and 0.08 of rgb2.center, anchor=center, image] (depth2){\includegraphics[width=\inputimagesize]{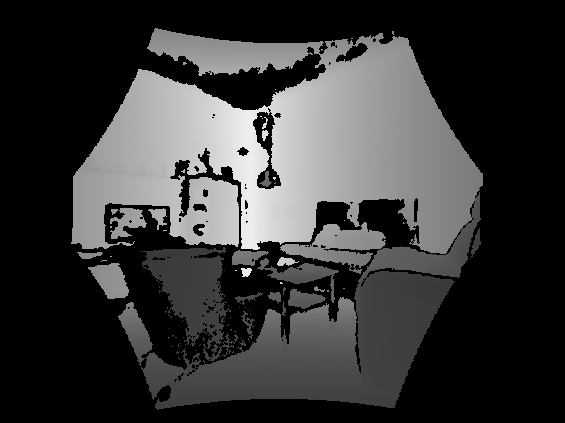}};%
        \end{scope}%

        \draw let%
            \p1=($(rgb2.north)!0.5!(depth2.south)$),%
            \p2=($(depth2.east)+(0.65, 0)$)%
        in%
            coordinate (encoder_anchor) at (\x2, \y1);%
        \node[emsaformer_encoder, anchor=south, font=\scriptsize, align=center] (encoder) at (encoder_anchor) {\rotatebox{90}{\parbox[c]{1cm}{\centering RGB-D Encoder}}};%
        \node[right= 0.1 of encoder.north, anchor=north, emsaformer_decoder, font=\scriptsize] (decoder) {\rotatebox{-90}{Decoder}};%

        \draw let%
            \p1=($(encoder.south)!0.5!(decoder.south)$)%
        in%
            node[gray] at ($(\p1)+(0, -0.8)$) {\scriptsize DVEFormer};%

        \begin{scope}[on background layer]%
            \fill[dveformer_box] ($(encoder.south east)-(0.2, 0.25)$) rectangle ($(decoder.south east)+(0.2, 0.45)$);%
        \end{scope}%

        \volume[right=0.65 of decoder.south, shift={(0, -0.1)}](output2)(1.3cm, 1.8cm, 0.1, 0.2);%
        \draw[-{Triangle[width=2.2cm,length=8pt]}, line width=1.7cm, application_arrow_color] ([shift={(0.1, -0.18)}]output2.south) -- ++(0, -0.6);%
        \node[below= 0.15 of output2, font=\scriptsize, shift={(0.1, 0)}, opacity=0.6] {Application};%

        \draw[-{Triangle[width=1cm,length=5pt]}, line width=0.5cm, data_arrow_color] ($(encoder.south)-(0.55, 0)$) -- ($(encoder.south)-(0.25, 0)$);%
        \draw[-{Triangle[width=1cm,length=5pt]}, line width=0.5cm, data_arrow_color] ($(decoder.south)+(0.25, 0)$) -- ($(decoder.south)+(0.55, 0)$);%

        \node[scale=0.7, font=\scriptsize, align=center] at (output2) {Dense\\Visual Embeddings};%
        \node[gray, font=\scriptsize, scale=0.7, right= 0.1 of output2, shift={(0, 0.1)}] {\rotatebox{90}{$H\times W\times D$}};%

        \node[left=0 of ndt_example_1, font=\scriptsize, gray] (ndtmaplabel) {\rotatebox{90}{3D NDT Map}};%
        \draw let%
            \p1=(ndtmaplabel),%
            \p2=(img_example_1.west)%
        in%
            node[font=\scriptsize, gray] (framelabel) at (\x1, \y2) {\rotatebox{90}{Frame-based\vphantom{p}}};%

        \begin{scope}[on background layer]%
            \fill[framebased_color] let%
                \p1=($(img_example_1.south west)-(0, 0.05)$),%
                \p2=(framelabel.south west),%
                \p3=($(ndt_example_3.north east)+(0.1, 0)$),%
                \p4=($(img_example_3.north)+(0, 0.1)$)%
            in%
                (\x2, \y1) rectangle (\x3, \y4);%
        \end{scope}%

        \path let%
            \p1=($(ndt_example_1)!0.5!(ndt_example_2)$),
            \p2=($(img_example_1.north)+(0, 0.4)$)
        in%
            coordinate (line_bottom) at (\x1, 0)%
            coordinate (line_top) at (\x1, \y2);%

        \draw[dashed] (line_bottom) -- (line_top);%
        
        \node[anchor=south west, left= 0.2 of line_top, font=\scriptsize, inner sep=0px, align=right] {Classical\\[-0.5mm]Semantic Segmentation};%
        \node[anchor=south east, right= 0.2 of line_top, font=\scriptsize, inner sep=0px, align=left] {\phantom{Semantic Segmentation}\\Text-based Querying};%

        \node[anchor=south west, chair_color, inner sep=1.5px, fill=framebased_color, fill opacity=0.8, text opacity=1, shift={(-0.001, -0.001)}] at (img_example_1.south west){%
            \tiny chair%
        };%
        \node[anchor=south west, chair_color, inner sep=1.5px, fill=white, fill opacity=0.8, text opacity=1, shift={(-0.001, -0.001)}] at (ndt_example_1.south west) {%
            \tiny chair%
        };%
    \end{tikzpicture}%
    \vspace{-1.5mm}%
    \caption{%
        A real-world example of our proposed approach, called DVEFormer, predicting dense visual embeddings~(DVE) on previously unseen data in one of our applications~\cite{Cohumanics-Fischedick-ISR-2023}. 
        Our approach not only supports classical semantic segmentation~(fixed predefined classes~--~closed set) but also enables text-based queries to specify objects of interest beyond a fixed spectrum of predefined classes.
        Moreover, as shown on the bottom, the predicted embeddings can be easily integrated into existing mapping approaches, such as~\cite{panopticndt2023iros}.
        Semantic colors are chosen as in~\cite{emsanet2022ijcnn} and are \href{https://github.com/TUI-NICR/nicr-scene-analysis-datasets/blob/v0.5.3/nicr_scene_analysis_datasets/datasets/nyuv2/nyuv2.py\#L193}{the default colors for NYUv2}~\cite{NYUv2-eccv2012}. 
        We refer to Fig.~\ref{fig:t-sne} for the subset of depicted semantic colors.
        Results for text-based queries show cosine similarity~(red: high, white: low).
    }%
    \label{fig:eye-catcher}%
    \vspace{-4mm}%
\end{figure}
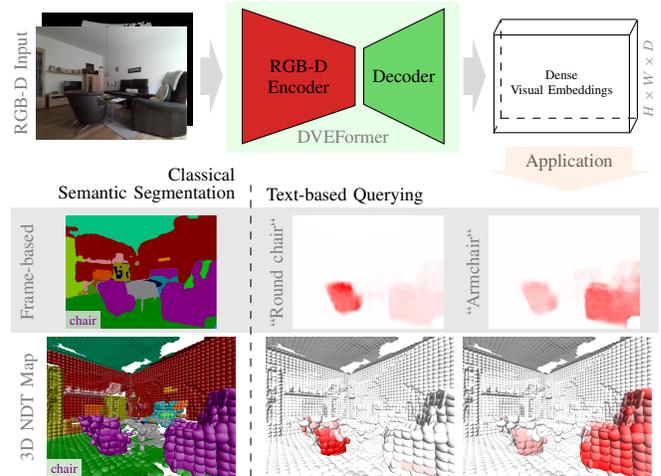%

However, image segmentation~\cite{DeepLabv3plus-eccv-2018, shapeconv-iccv2021, esanet2021icra, mulitmae-cvpr2022, omnivore-cvpr2022, emsanet2022ijcnn, emsaformer2023ijcnn, cmx-its2023, cmnext-cvpr2023, dformer-iclr2024} and 3D mapping approaches~\cite{semanticmapping2022icra, panopticndt2023iros, panoptic-fusion-2019-iros, panoptic-multi-tsdf-2022-icra,semantic-bsk-map-RAL-2020, oneformer3d-cvpr2023, scalable_panoptic_3dv2024} typically only operate on a fixed class spectrum that might be insufficient for many real-world applications.
This is because they typically rely on the class spectrum of common indoor segmentation datasets, such as NYUv2~\cite{NYUv2-eccv2012}, SUN RGB-D~\cite{SUNRGBD-cvpr2015}, or ScanNet~\cite{scannet-cvpr2017}, which only provide the classes \textit{chair} and \textit{sofa} for the given example.
While there are already approaches for open-vocabulary segmentation~\cite{maskclip-eccv2022, zegformer-cpvr2022, sed-cpvr2024, cat-seg-cvpr2024, zero-shot-neurips2019, semantic-projection-cvpr2019, openvocab-cvpr2023, maft-neurips2023, openvocab-cvpr2024, multi-view-open-vocab-2023prcv}, they are often unsuitable for mobile applications with limited computational resources.
These approaches are increasingly complex, require multiple large backbones, or are not specifically tailored for indoor settings.

To overcome these limitations, we propose Dense Visual Embedding Transformer~(DVEFormer)~--~a knowledge distillation approach~\cite{hinton2015distilling} for dense prediction of text-aligned visual embeddings using an efficient Transformer-based~\cite{Transformer-neurips2017} architecture.
Instead of relying on the predefined semantic classes of the used indoor dataset, we only use their provided mask annotations~--~ without their semantic label~--~as input to Alpha-CLIP~\cite{alphaclip2024cvpr} to extract teacher embeddings.
Our student model~--~DVEFormer~--~is based on the EMSAFormer architecture~\cite{emsaformer2023ijcnn}, featuring a modified Swin-Transformer-based~\cite{swinv2-cvpr2022} RGB-D encoder and a lightweight SegFormer-like~\cite{SegFromer-neurips2021} decoder to predict dense segmentations, while being optimized for efficient inference on mobile robots.
Instead of predicting predefined semantic classes, we train our model to directly output dense visual embeddings distilled from the teacher.
This way, our predicted embeddings are text-aligned and can be queried at runtime by comparing them with incoming text embeddings without retraining the model itself.
To further highlight the practical application of our approach, we show qualitative results for directly integrating the estimated embeddings into a 3D mapping pipeline~\cite{Einhorn-ECMR-2013-NDT, semanticmapping2022icra, panopticndt2023iros}, resulting in a comprehensive 3D scene representation.
This representation can then be queried by untrained operators with text to, for example, specify targets for robot navigation.

We share the pipeline for data preparation as well as the training details and model weights at GitHub: \href{https://github.com/TUI-NICR/DVEFormer}{\small\texttt{\url{https://github.com/TUI-NICR/DVEFormer}}}.

\section{Related Work}
\label{sec:related_work}
In the following, we summarize related work.
We begin by reviewing methods for dense RGB-D semantic segmentation.
Next, we discuss vision–language models that align visual and textual modalities, highlighting their ability to produce global representations while noting their limitations for fine-grained tasks and mobile platforms. 
Finally, we briefly discuss knowledge distillation as a method to transfer features from large teacher models to more efficient student models.
\subsection{RGB-D Semantic Segmentation}
Dense semantic segmentation can be done in different ways. 
Most methods involve pixel-wise predictions through an encoder–decoder approach~\cite{DeepLabv3plus-eccv-2018, emsanet2022ijcnn, emsaformer2023ijcnn, cmx-its2023, cmnext-cvpr2023, dformer-iclr2024, mulitmae-cvpr2022}, whereas recent methods, such as \cite{Max-Deeplab-cvpr2021, maskformer-neurips2021, mask2former-cvpr2022, panoptic-segformer-cvpr2022}, predict larger segments assigned to fixed classes through a Transformer-based decoder.
Although these mask-wise approaches have demonstrated strong performance, they tend to be complex and are rarely optimized for mobile applications, which is why we focus on pixel-wise predictions.

Combining RGB and depth data is an effective strategy to improve segmentation performance, as both modalities provide complementary information~\cite{FuseNet-accv2016, RedNet-arxiv2018, ACNet-icip2019}.
RGB images offer semantic cues such as color and texture, while depth images provide geometric information.
To combine RGB and depth, some approaches~\cite{SSMA-ijcv2019, SA-Gate-eccv2020,emsanet2022ijcnn, FuseNet-accv2016} process both modalities separately using two or more encoders and fuse features at different stages.
While this strategy can be effective, it requires carefully designing the fusion mechanism and often leads to increased computational overhead, depending on the used backbone and fusion mechanism.
An alternative approach is to use a single encoder that combines RGB and depth as joint input to a model. 
Early CNN-based approaches achieve this by employing special depth-aware convolutions~\cite{DepthAwareCNN-eccv2018, Malleable2.5D-eccv2020, shapeconv-iccv2021}.
However, such operations are often not suitable for mobile platforms.
In addition to these methods, Transformer-based approaches have been proposed.
Some of these methods employ Vision Transformers~\cite{vit-iclr2021} to process both modalities in a single transformer without a pyramidal structure~\cite{mulitmae-cvpr2022}, while others use Transformer-based encoders that follow a pyramid structure common in CNNs, making them suitable as drop-in replacements in existing models~\cite{omnivore-cvpr2022, emsaformer2023ijcnn, dformer-iclr2024}.

In our work, we build upon EMSAFormer~\cite{emsaformer2023ijcnn}, which uses a single modified Swin-V2-T encoder to jointly process RGB and depth data.
Compared to the original Swin-V2-T backbone, only minor modifications are made for depth integration, and no special operations are introduced that might be unsupported or suboptimal for inference engines such as NVIDIA TensorRT.

\subsection{Vision–Language Models for Dense Predictions}%
Vision–language models are highly valued as they enable the alignment of visual and textual modalities in a shared embedding space.
CLIP~\cite{clip-icml2021} was a breakthrough in this area by jointly training a vision encoder and a text encoder through contrastive learning, achieving impressive zero-shot performance on various image classification tasks.
However, CLIP produces only a single global embedding for an entire image, which limits its capabilities for tasks that require fine-grained localized embeddings.

To overcome this limitation, several approaches have been developed to generate more detailed embeddings.
One strategy involves extracting local embeddings by cropping around objects of interest~\cite{reclip-2022etal, ovarnet-2023cpvr}.
Approaches such as MaskCLIP~\cite{maskclip-eccv2022} and CAT-Seg~\cite{cat-seg-cvpr2024} perform semantic segmentation in both closed-set and open-vocabulary settings by either comparing local visual embeddings directly to text embeddings or incorporating additional layers for dense prediction.
Other approaches~\cite{zegformer-cpvr2022, openvocab-cvpr2023, openvocab-cvpr2024, multi-view-open-vocab-2023prcv} also introduce additional mechanisms to derive local embeddings from a global CLIP embedding.
However, these approaches often require the simultaneous use of two large vision encoders and, thus, increase computational complexity. 
By contrast, Alpha-CLIP~\cite{alphaclip2024cvpr} incorporates an additional alpha channel as an input to guide the model in generating embeddings for specific regions without altering the original image content.
Although this approach requires multiple inferences per image if different masks are of interest, it preserves valuable contextual information and retains the generalization performance of the original CLIP model.
To further improve segmentation performance, some methods~\cite{maskclip-eccv2022, ovarnet-2023cpvr, maft-neurips2023, multi-view-open-vocab-2023prcv} integrate a knowledge distillation or pseudo label-based strategy.
For example, MaskCLIP+~\cite{maskclip-eccv2022} uses predictions from MaskCLIP as ground-truth labels to guide the training of a segmentation-specific model. 
This approach overcomes the architectural rigidity of directly using the CLIP image encoder and improves performance compared to MaskCLIP.

In our work, we leverage Alpha-CLIP to generate teacher embeddings that serve as targets in a knowledge distillation framework, enabling our student model DVEFormer to predict dense pixel-wise embeddings that are aligned with the vision–language space of Alpha-CLIP and CLIP.

\newcommand{\imagewidth}{2cm}%
\newcommand{\figheight}{8}%
\newcommand{\offlineareawidth}{0.37\hsize}%
\newcommand{\onlineareawidth}{0.62\hsize}%
\newcommand{\sectionmargin}{0.1}%
\definecolor{alphaclip_color}{HTML}{009900}%
\definecolor{textembedder_color}{HTML}{009900}%
\definecolor{emsaformer_encoder_color}{HTML}{d62727}%
\definecolor{emsaformer_context_color}{HTML}{ffc266}%
\definecolor{emsaformer_decoder_color}{HTML}{6cd56a}%
\definecolor{emsaformer_upsampling_color}{HTML}{37967A}%
\definecolor{offline_color}{HTML}{ff5f6b}%
\definecolor{model_color}{HTML}{6bff5f}%
\definecolor{train_color}{HTML}{a35fff}%
\definecolor{evaluation_color}{HTML}{ffa35f}%
\definecolor{mask_0}{HTML}{f8b973}%
\definecolor{mask_1}{HTML}{e9516b}%
\definecolor{mask_2}{HTML}{007fff}%
\definecolor{mask_3}{RGB}{64, 0, 0}%
\definecolor{mask_4}{RGB}{64, 128, 0}%
\definecolor{mask_5}{RGB}{128, 0, 0}%
\definecolor{area_0}{RGB}{239, 127, 216}%
\definecolor{area_1}{RGB}{56, 1, 254}%
\definecolor{area_2}{RGB}{88, 251, 192}%
\definecolor{area_3}{RGB}{194, 249, 49}%
\definecolor{area_4}{RGB}{131, 155, 48}%
\definecolor{area_5}{RGB}{0, 127, 255}%
\def\embvector[#1](#2)(#3, #4){
    \node[embeddingvector, #1] (#2) {\rotatebox{90}{\scriptsize ...}};
    \node[anchor=north, inner sep=0, shift={(0, -0.1)}] at (#2.north) {\tiny #3};
    \node[anchor=south, inner sep=0, shift={(0, 0.1)}] at (#2.south) {\tiny #4};
    \foreach \frac in {0.3, 0.7}
    {
        \draw[dotted] ($(#2.south west)!\frac!(#2.north west)$) -- ($(#2.south east)!\frac!(#2.north east)$);
    }
}%
\begin{figure*}[!b]%
    \vspace{-2mm}%
    \centering%
    \tikzset{alphaclip/.style={fill=alphaclip_color, font=\scriptsize, align=center, inner sep=2px, draw}}%
    \tikzset{textembedder/.style={fill=textembedder_color, font=\scriptsize, align=center, inner sep=2px, draw}}%
    \tikzset{emsaformer_encoder/.style={fill=emsaformer_encoder_color, minimum width=3cm, minimum height=1.3cm, trapezium, trapezium angle=40, rotate=-90, draw}}
    \tikzset{emsaformer_decoder/.style={fill=emsaformer_decoder_color, minimum width=3cm, minimum height=0.8cm, trapezium, trapezium angle=40, rotate=90, draw}}
    \tikzset{emsaformer_context/.style={fill=emsaformer_context_color, font=\scriptsize, draw, inner sep=3px}}
    \tikzset{dataarrow/.style={arrows = {-Stealth}, black, thick}}
    \tikzset{embeddingvector/.style={fill=white, draw, minimum width=0.3cm, minimum height=1cm}}
    \tikzset{image/.style={inner sep=0, line width=1.5px}}
    \tikzset{section/.style={line width=0px, opacity=0.15, inner sep=0}}
    \tikzset{textprompt/.style={font=\scriptsize, align=center, scale=0.8}}
    \tikzset{sectionlabel/.style={font=\footnotesize, gray, inner sep=2px}}
    \tikzset{datalabel/.style={font=\small}}
    \tikzset{datalabelsmall/.style={font=\scriptsize}}
    \tikzset{shapeannotation/.style={font=\scriptsize, scale=0.7, gray, rotate=90}}
    \begin{tikzpicture}
        \tikzset{trapezium stretches=true}

        \coordinate (alphaclip_anchor) at (3.9, 0);
        \coordinate (top_right) at (0.99\hsize, 7.55);
        \coordinate (center_left) at (0, 3.25);
        \path let
            \p1=($(top_right)!0.5!(center_left)$)
        in 
            coordinate[shift={(1.2, 0)}] (emsaformer_anchor) at (\offlineareawidth, \y1);
        \path let
            \p1=(center_left)
        in
            coordinate (training_bottom_left) at (14, \y1);

        \fill[train_color, section] let
            \p1=(training_bottom_left),
            \p2=(top_right)
        in
            (\p1) rectangle (\p2);
        \node[anchor=south west, sectionlabel] at (training_bottom_left) {\small Training (Sec.~\ref{sec:model_and_knowledge_distillation})};

        \fill[evaluation_color, section] let
            \p1=($(center_left)-(0, \sectionmargin)$),
            \p2=(top_right)
        in
            (0, 0) rectangle (\x2, \y1);
        \node[sectionlabel, anchor=south west] at (0, 0) {Application (Sec.~\ref{sec:model_interpret_embedding})};

        \fill[model_color, section] let
            \p1=($(\offlineareawidth, 0)+(\sectionmargin, 0)$),
            \p2=(center_left),
            \p3=($(training_bottom_left)-(\sectionmargin, 0)$),
            \p4=(top_right)
        in
            (\x1, \y2) rectangle (\x3, \y4);
        \draw let
            \p1=($(\offlineareawidth, 0)+(\sectionmargin, 0)$),
            \p2=(center_left),
        in
            node[anchor=south west, sectionlabel] at (\x1, \y2) {DVEFormer\vphantom{p} (Sec.~\ref{sec:model_and_knowledge_distillation})};

        \fill[section, offline_color] let
                \p1=(top_right),
                \p2=(center_left)
        in
            (0, \y2) rectangle (\offlineareawidth, \y1);
        \node[anchor=south west, sectionlabel] at (center_left) {Dataset Preparation (Sec.~\ref{sec:embedding_vector_estimation})};
        
        \begin{scope}[name=data_instance]

            \draw let
                \p1=(emsaformer_anchor)
            in
                node[anchor=north west, image, shift={(0, -0.1)}] (rgb) at (0.5, \y1) {
                    \includegraphics[width=\imagewidth]{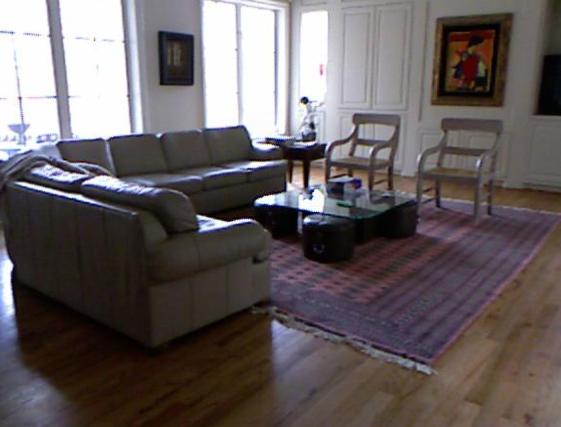}
                };
            \node[left= 0 of rgb] {\rotatebox{90}{$\sIRGB$}};
            
            \node[above=0.2 of rgb.north, image, draw=mask_2, shift={(0.1, 0.1)}] (alpha_anchor) {
                \includegraphics[width=\imagewidth]{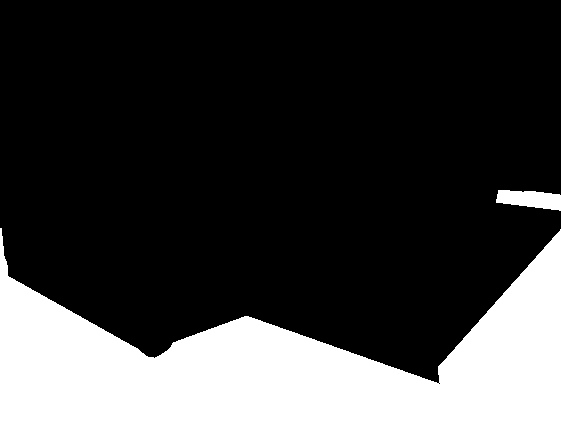}
            };
            \node[image, draw=mask_1, shift={(-0.1, -0.1)}] (alpha) at (alpha_anchor) {
                \includegraphics[width=\imagewidth]{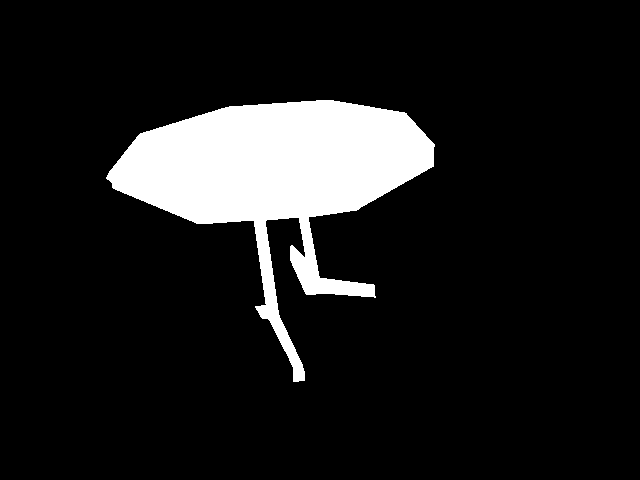}
            };
            \node[image, draw=mask_0, shift={(-0.2, -0.2)}] at (alpha_anchor) {
                \includegraphics[width=\imagewidth]{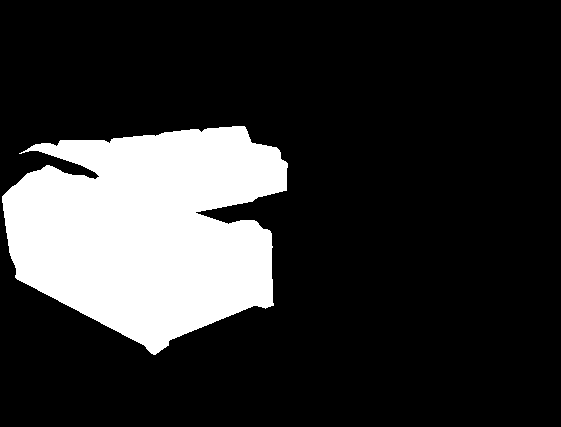}
            };

            \node[left= 0 of alpha] {\rotatebox{90}{$\setM\vphantom{_RGB}$}};

            \draw let 
                \p1=(rgb),
                \p2=(alphaclip_anchor)
            in
                node[alphaclip] (alphaclip_scene) at (\x2, \y1) {Alpha-CLIP\\Vision-Encoder};

            \embvector[draw, right= 0.5 of alphaclip_scene](embedding_scene)(-0.3, -0.2);
            \node[datalabelsmall, below=0.05 of embedding_scene] {$\seI$};
 
            \draw[dataarrow] (alphaclip_scene) -- (embedding_scene);

            \draw let \p1=(alpha), \p2=(alphaclip_scene) in node[alphaclip, shift={(0, -0.1)}] (alphaclip) at (\x2, \y1) {Alpha-CLIP\\Vision-Encoder};

            \draw[dataarrow] (alpha.east) -- ++ (0.2, 0) |- ([shift={(0, 0.1)}]alphaclip.west);
            \draw[dataarrow] (rgb.east) -- ++ (0.2, 0) |- ([shift={(0, -0.1)}]alphaclip.west);
            \draw[dataarrow] (rgb.east) -- ++ (0.2, 0) |- (alphaclip_scene.west);

            \embvector[draw=mask_2, right= 0.4 of alphaclip, shift={(0.1, 0.1)}](emb_0)(-0.3, -0.1);
            \embvector[draw=mask_1, right= 0.4 of alphaclip](embedding_instance)(-0.2, -0.1);
            \embvector[draw=mask_0, right= 0.4 of alphaclip, shift={(-0.1, -0.1)}](emb_2)(-0.4, 0.1);

            \node[datalabelsmall, below=0.05 of embedding_instance] {$\{\seMi\}$};

            \draw(embedding_instance)++(0.76,0) circle[radius=5px] node (sub) {\small $-$};
            \draw (sub)++(0, -1) circle[radius=5px] node (alpha_scale) {\small $\cdot\alpha$};
            \draw[dataarrow] (alphaclip) -- ([shift={(-0.1, 0)}]embedding_instance.west);
            \draw[dataarrow] ([shift={(0.1, 0)}]embedding_instance.east) -- (sub);

            \draw[dataarrow] (embedding_scene) -| (alpha_scale);
            \draw[dataarrow] (alpha_scale) -- (sub);

            \node[datalabel, above= 0.5 of sub] (data_final_embs) {$\{\steMi\}$};
            \draw[dataarrow] (sub) -- (data_final_embs);
        \end{scope}

        \begin{scope}[name=emsaformer, shift={(\offlineareawidth, 0)}]

            \node[image, below= 0.1 of emsaformer_anchor] (depth) {
                \includegraphics[width=\imagewidth]{img/dataset/approach/examples/kv1_NYUdata_02428/rgb.png}
            };
            \node[above=0.1 of emsaformer_anchor, image] (rgb) {
                \includegraphics[width=\imagewidth]{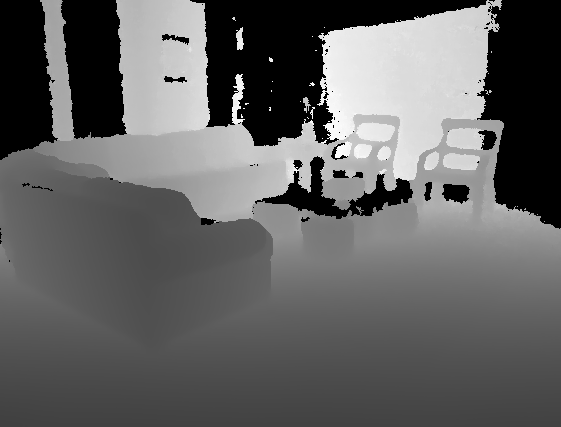}
            };

            \draw let \p1=(emsaformer_anchor) in node[emsaformer_encoder] (emsaformer_encoder) at (3.5, \y1) {\rotatebox{180}{\small Encoder}};

            \node[right= 0.3 of emsaformer_encoder.north, emsaformer_context] (emsaformer_context) {
                \begin{tikzpicture}
                    \node[rotate=90] {Context Module};
                \end{tikzpicture}
            };
            
            \node[right=0.3 of emsaformer_context.east, anchor=north, emsaformer_decoder] (emsaformer_decoder) {\small Decoder};

            \node[right= 0.3 of emsaformer_decoder.south, font=\scriptsize, draw, fill=emsaformer_upsampling_color, dashed, fill opacity=0.7, text opacity=1] (upsampling) {\rotatebox{90}{Optional Upsampling $\times 4$}};

            \draw[dataarrow] (depth.east) -- ++ (0.1, 0) |- (emsaformer_encoder.south);
            \draw[dataarrow] (rgb.east) -- ++ (0.1, 0) |- (emsaformer_encoder.south);

            \draw[dataarrow] (emsaformer_encoder.north) -- (emsaformer_context.west) node[shapeannotation, above, shift={(-0.45, -0.03)}] {$H/32\times W/32 ~~~\times 1024$};

            \draw[dataarrow] (emsaformer_context.east) -- (emsaformer_decoder.north) node[shapeannotation, above, shift={(-0.45, -0.03)}] {$H/32\times W/32 ~~~\times 1024$};
            \draw[dataarrow] (emsaformer_decoder.south) -- (upsampling.west) node[shapeannotation, above, shift={(-0.35, -0.03)}] {$H/4\times W/4 ~~~\times 768$};

            \node[fill=orange, right=0.7 of upsampling, align=center, inner sep=1, draw] (loss) {
            \begin{tikzpicture}
                    \node[rotate=90, align=center, font=\scriptsize] {Cosine Distance\\Loss};
                \end{tikzpicture}
            };

            \draw[dataarrow] (upsampling.east)  node[shapeannotation, above, shift={(-0.1, -0.425)}] {$H\times W ~~~\times 768$} -- (loss);
            
            \node[right = 0.3 of loss, image, line width=0px, shift={(0, -0.2)}] (embedding_image) {
                \includegraphics[width=\imagewidth]{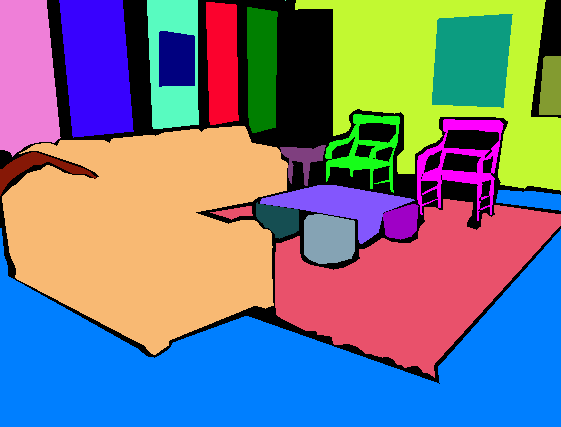}
            };
            \draw (embedding_image.south east) -- ++(0.2, 0.4);
            \draw[dashed] (embedding_image.south east)++(0, 0.4) -- ++(0.2, 0);
            \draw (embedding_image.north west) -- ++(0.2, 0.4);
            \draw[dashed] (embedding_image.north west)++(0.2, 0) -- ++(0, 0.4);
            \draw (embedding_image.north east) -- ++(0.2, 0.4);
            \draw let 
                \p1=($(embedding_image.west)-(embedding_image.east)$),
                \n1 = {veclen(\p1)-\pgflinewidth}
            in 
                (embedding_image.north west)++(0.2, 0.4) -- ++(\n1, 0);

            \draw let 
                \p1=($(embedding_image.north)-(embedding_image.south)$),
                \n1 = {veclen(\p1)-\pgflinewidth}
            in 
                (embedding_image.south east)++(0.2, 0.4) -- ++(0, \n1);

            \node[datalabel, below=0.1 of embedding_image] {$\sY$};

            \foreach \fraca/\fracb/\c in {0/0.1/area_0, 0.1/0.25/area_1, 0.25/0.35/area_2, 0.35/0.53/black, 0.53/0.97/area_3, 0.97/1/black} {
                \draw[opacity=0.8, densely dotted] ($(embedding_image.north west)!\fracb!(embedding_image.north east)$) -- ++(0.2, 0.4);
                \fill[opacity=0.6, \c] let
                    \p1=($(embedding_image.north west)!\fraca!(embedding_image.north east)$),
                    \p2=($(embedding_image.north west)!\fracb!(embedding_image.north east)$),
                    \p3=($(\p1) - (\p2)$),
                    \n1={veclen(\p3)}
                in
                    (\p1) -- (\p2) -- ++(0.2, 0.4) -- ++(-\n1,  0) -- cycle;
            };
            \foreach \fraca/\fracb/\c in {0/0.13/black, 0.13/0.26/area_4, 0.26/0.45/area_3, 0.45/1/area_5} {
                \draw[opacity=0.8, densely dotted] ($(embedding_image.north east)!\fracb!(embedding_image.south east)$) -- ++(0.2, 0.4);
                \fill[opacity=0.6, \c] let
                    \p1=($(embedding_image.north east)!\fraca!(embedding_image.south east)$),
                    \p2=($(embedding_image.north east)!\fracb!(embedding_image.south east)$),
                    \p3=($(\p1) - (\p2)$),
                    \n1={veclen(\p3)}
                in
                    (\p1) -- (\p2) -- ++(0.2, 0.4) -- ++(0, \n1) -- cycle;
            };
            
            \node[draw, fill=orange, above= 0.7 of embedding_image, align=center, minimum height=0.8cm, shift={(0.1, 0)}, inner sep=1, font=\scriptsize] (emb_to_img) {Embeddings\\to Teacher};

            \node[datalabel, left=0.5 of emb_to_img, shift={(0, 0.2)}] (train_masks) {$\setM$};
            \node[datalabel, left=0.5 of emb_to_img, shift={(0, -0.2)}] (train_embs) {$\{\steMi\}$};
            \draw[dataarrow] (train_masks) -- ([shift={(0, 0.2)}]emb_to_img.west);
            \draw[dataarrow] (train_embs) -- ([shift={(0, -0.2)}]emb_to_img.west);
            
            \draw[dataarrow] (emb_to_img) -- ([shift={(0.1, 0.4)}]embedding_image.north);

            \draw[dataarrow] ([shift={(0, 0.2)}]embedding_image.west) -- (loss);

        \end{scope}

        \begin{scope}[name=inference, shift={(\offlineareawidth, 0)}]
            \node[textembedder, above=2.15 of alphaclip_anchor] (textembedder_cs) {Alpha-CLIP\\Text-Encoder};

            \node[textprompt, left=0.5 of textembedder_cs, font=\scriptsize] (prompts) {
                ``A picture of a chair``\\
                \rotatebox{90}{...}\\
                ``A picture of a wall``
            };

            \embvector[right=0.5 of textembedder_cs, shift={(0.1, 0.1)}](text_emb0)(-0.3, 0.2);
            \embvector[right=0.5 of textembedder_cs](text_emb1)(0, 0);
            \embvector[right=0.5 of textembedder_cs, shift={(-0.1, -0.1)}](text_emb2)(-0.3, 0.0);

            \draw[dataarrow] (prompts) -- (textembedder_cs);
            \draw[dataarrow] (textembedder_cs) -- ([shift={(-0.1, 0)}]text_emb1.west);

            \node[textembedder, above=0.9 of alphaclip_anchor] (textembedder_os) {Alpha-CLIP\\Text-Encoder};

            \node[textprompt, left=0.5 of textembedder_os, font=\scriptsize] (prompt) {
                ``Orange picture``
            };

            \embvector[right=0.5 of textembedder_os](text_emb)(0.1, 0.3);

            \draw[dataarrow] (prompt) -- (textembedder_os);
            \draw[dataarrow] (textembedder_os) -- (text_emb.west);

            \phantomsubcaption\label{fig:approach::application_querry}%
            \phantomsubcaption\label{fig:approach::application_text_based}%
            \phantomsubcaption\label{fig:approach::application_visual_mean}%
            \phantomsubcaption\label{fig:approach::application_linear_probing}%

            \draw let
                \p1=(textembedder_os)
            in
                node[image,draw, line width=0.5px] at (0.35*\onlineareawidth, \y1) (inference_seg) {
                    \includegraphics[width=\imagewidth]{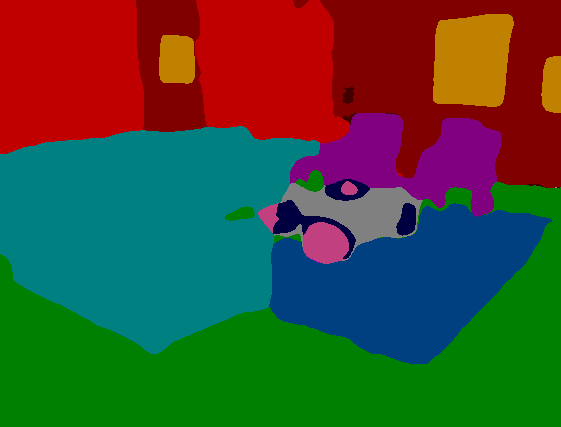}
                };
            \node[below=-0.05 of inference_seg, align=center, font=\scriptsize] {\subref*{fig:approach::application_text_based} Text-based};

            \node[image, left=0.7 of inference_seg, draw, line width=0.5px] (inference_query) {
                \includegraphics[width=\imagewidth, trim={0, 0, 4px, 3px}, clip]{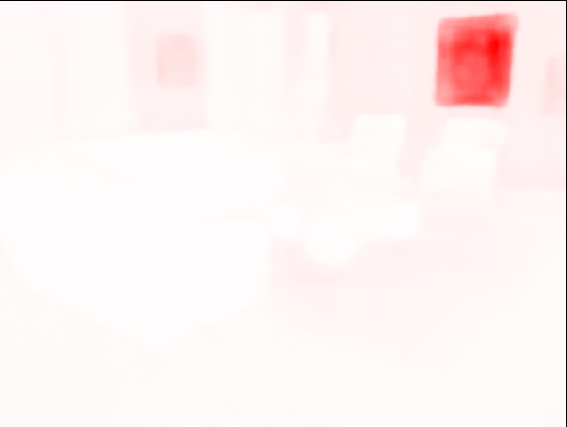}
            };
            \node[below=-0.05 of inference_query, font=\scriptsize] {\subref*{fig:approach::application_querry} Query};

            \node[image, right=0.7 of inference_seg, draw, line width=0.5px] (inference_seg_vm) {
                \includegraphics[width=\imagewidth]{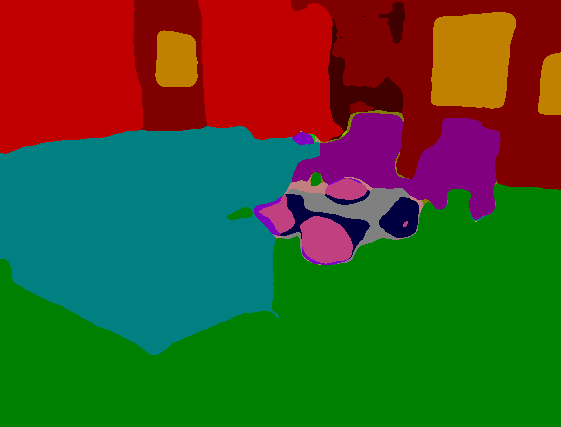}
            };
            \node[below=-0.05 of inference_seg_vm, align=center, font=\scriptsize] {\subref*{fig:approach::application_visual_mean} Visual Mean};
            \node[image, right=0.7 of inference_seg_vm, draw, line width=0.5px] (inference_seg_prob) {
                \includegraphics[width=\imagewidth]{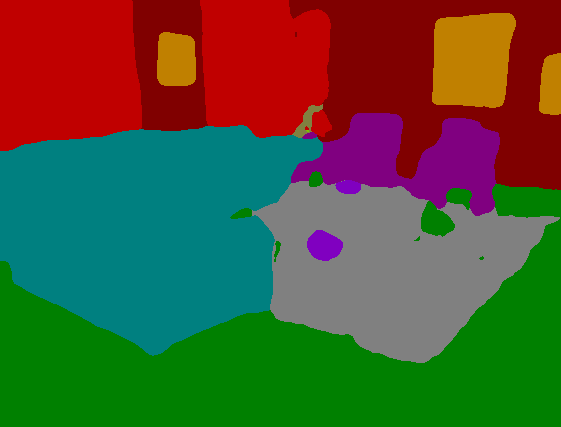}
            };
            \node[below=-0.05 of inference_seg_prob, align=center, font=\scriptsize] {\subref*{fig:approach::application_linear_probing} Linear Probing};

            \draw[dataarrow] let
                \p1=($(text_emb1)+(0, 0.4)$),
                \p2=($(upsampling.east)+(0.3, 0)$),
                \p3=($(inference_seg.north)+(0.5, 0)$)
            in
                (upsampling.east) -- (\p2) -- (\x2, \y1) -| (\p3);
            \draw[dataarrow] let
                \p1=($(text_emb1)+(0, 0.4)$),
                \p2=($(upsampling.east)+(0.3, 0)$),
                \p3=($(inference_query.north)$)
            in
                (upsampling.east) -- (\p2) -- (\x2, \y1) -| (\p3);
            \draw[dataarrow] let
                \p1=($(text_emb1)+(0, 0.4)$),
                \p2=($(upsampling.east)+(0.3, 0)$),
                \p3=($(inference_seg_prob.north)+(0.5, 0)$)
            in
                (upsampling.east) -- (\p2) -- (\x2, \y1) -| (\p3);
            \draw[dataarrow] let
                \p1=($(text_emb1)+(0, 0.4)$),
                \p2=($(upsampling.east)+(0.3, 0)$),
                \p3=($(inference_seg_vm.north)+(0.5, 0)$)
            in
                (upsampling.east) -- (\p2) -- (\x2, \y1) -| (\p3);
                
            \draw[dataarrow, dashed] (text_emb.east) -- ++(1, 0) |- (inference_query);
            \draw[dataarrow, dashed] ([shift={(0.1, 0)}]text_emb1.east) -| ([shift={(-0.5, 0)}]inference_seg.north);

            \draw[dataarrow, dashed] let
                \p1=(text_emb1),
                \p2=(embedding_image.south east),
                \p3=($(inference_seg_vm.north)-(0.5, 0)$)
            in
                (\p2) -- ++(0, 0) -- ++(0, \y1-\y2) -| (\p3);
            \draw[dataarrow, dashed] let
                \p1=(text_emb1),
                \p2=(embedding_image.south east),
                \p3=($(inference_seg_prob.north)-(0.5, 0)$)
            in
                (\p2) -- ++(0, 0) -- ++(0, \y1-\y2) -| (\p3);

        \end{scope}%
    \end{tikzpicture}%
    \vspace{-1.5mm}%
    \caption{%
        Overview of our proposed approach on an example image of SUN RGB-D~\cite{SUNRGBD-cvpr2015}. %
        Alpha‑CLIP is used offline to process RGB images and a set of binary segment masks to extract a set of teacher embeddings.
        These guide our efficient Dense Visual Embedding Transformer~(DVEFormer) to learn dense pixel‑wise visual embeddings via knowledge distillation.
        For downstream applications, the resulting embeddings can be used for text‑based retrieval \subref*{fig:approach::application_querry}~or classical semantic segmentation with predefined classes \subref*{fig:approach::application_text_based}-\subref*{fig:approach::application_linear_probing}, or other robotic applications, enabling flexible scene understanding tasks. %
        We refer to Fig.~\ref{fig:t-sne} for the subset of depicted semantic colors. %
    }%
    \label{fig:approach}%
\end{figure*}%

\subsection{Knowledge Distillation for Dense Visual Embeddings}%
Knowledge distillation~\cite{hinton2015distilling}, i.e., transferring knowledge from large computationally intensive models to smaller and more efficient ones, has been proven in many domains.
For example, in natural language processing, large language models are often distilled into compact models that retain most of the teacher's capabilities~\cite{distilbert-2019neurips, tinypert-2019arxiv}.
Distillation can be performed by various methods, e.g., by minimizing losses that compare the teacher's and student's outputs~\cite{distilbert-2019neurips, tinypert-2019arxiv, maskclip-eccv2022, ovarnet-2023cpvr}.
In computer vision, training models to predict robust visual embeddings like those provided by CLIP requires large amounts of data and significant computational resources, which can be impractical for dense, pixel-wise tasks.

We leverage existing datasets that provide binary segment masks and use Alpha-CLIP to generate teacher embeddings as pseudo ground-truth labels.
Our approach uses these teacher embeddings to supervise a student model, similar to what is done for MaskCLIP+, enabling it to predict dense, pixel-wise embeddings that are well aligned with the vision–language space while significantly reducing computational overhead during inference.

\section{Efficient Prediction of Dense Visual Embeddings}%
\label{sec:main}
Our approach is shown in Fig.~\ref{fig:approach} and starts by extracting teacher embeddings using Alpha-CLIP~\cite{alphaclip2024cvpr}, as explained in Sec.~\ref{sec:embedding_vector_estimation}.
The extracted embeddings are used to train our student model~--~Dense Visual Embedding Transformer (DVEFormer). 
We describe the model architecture and the distillation training in Sec.~\ref{sec:model_and_knowledge_distillation}.
DVEFormer predicts dense pixel-wise embeddings that encode comprehensive information for various downstream tasks. 
For example, they can be paired with an aligned text encoder for text-guided retrieval or used for classification tasks, as described in Sec.~\ref{sec:model_interpret_embedding}. 
\subsection{Segment-Level Visual Embedding Extraction}%
\label{sec:embedding_vector_estimation}%
Our method requires pixel-level visual embeddings as teacher information for offline knowledge distillation.
To extract them, for each dataset sample, the RGB image~$\sIRGB$ is paired  with each of the corresponding binary segment masks $\sMi \in \setM$.
The binary segment masks are derived from panoptic annotations provided in extended version of the datasets~\cite{emsanet2022ijcnn, emsaformer2023ijcnn}, which include both distinct objects~(e.g., chair or table) and uncountable regions (e.g., wall or floor).
Each pair~$(\sIRGB, \sMi)$ is processed by Alpha-CLIP to generate an embedding $\seMi \in \mathbb{R}^{D}$ with $D=768$.
While the dimensionality might be higher than required for our application, we retain the full dimensionality to ensure alignment with the existing text encoder used by Alpha-CLIP, avoiding potential artifacts from dimensionality reduction.

Fig.~\ref{fig:t-sne::sem} shows a t-SNE visualization~\cite{tsne-jmlr2008} of the extracted segment embeddings~$\seMi$ for the NYUv2 dataset~\cite{NYUv2-eccv2012}.
It reveals that they tend to cluster by the scene class rather than the semantic class of the segment.
\definecolor{highlight_color}{HTML}{800080}%
\definecolor{bathroom_color}{HTML}{777777}%
\definecolor{bedroom_color}{RGB}{244, 243, 131}%
\definecolor{diningroom_color}{RGB}{137, 28, 157}%
\definecolor{discussionroom_color}{RGB}{150, 255, 255}%
\definecolor{hallway_color}{RGB}{54, 114, 113}%
\definecolor{kitchen_color}{HTML}{0000B0}%
\definecolor{livingroom_color}{RGB}{255, 69, 0}%
\definecolor{office_color}{HTML}{5770FF}%
\definecolor{otherindoor_color}{HTML}{00a321}%
\definecolor{wall_color}{HTML}{800000}%
\definecolor{floor_color}{HTML}{008000}%
\definecolor{window_color}{HTML}{c00000}%
\definecolor{picture_color}{HTML}{c08000}%
\definecolor{sofa_color}{HTML}{008080}%
\definecolor{tv_color}{HTML}{c04000}%
\definecolor{ceiling_color}{HTML}{00c080}%
\definecolor{table_color}{HTML}{808080}%
\definecolor{cabinet_color}{HTML}{808000}%
\definecolor{box_color}{HTML}{c04080}%
\definecolor{lamp_color}{HTML}{808040}%
\definecolor{floormat_color}{HTML}{004080}%
\definecolor{otherprop_color}{HTML}{400040}%
\definecolor{otherstructure_color}{HTML}{0080C0}%
\definecolor{otherfurniture_color}{HTML}{8080c0}%
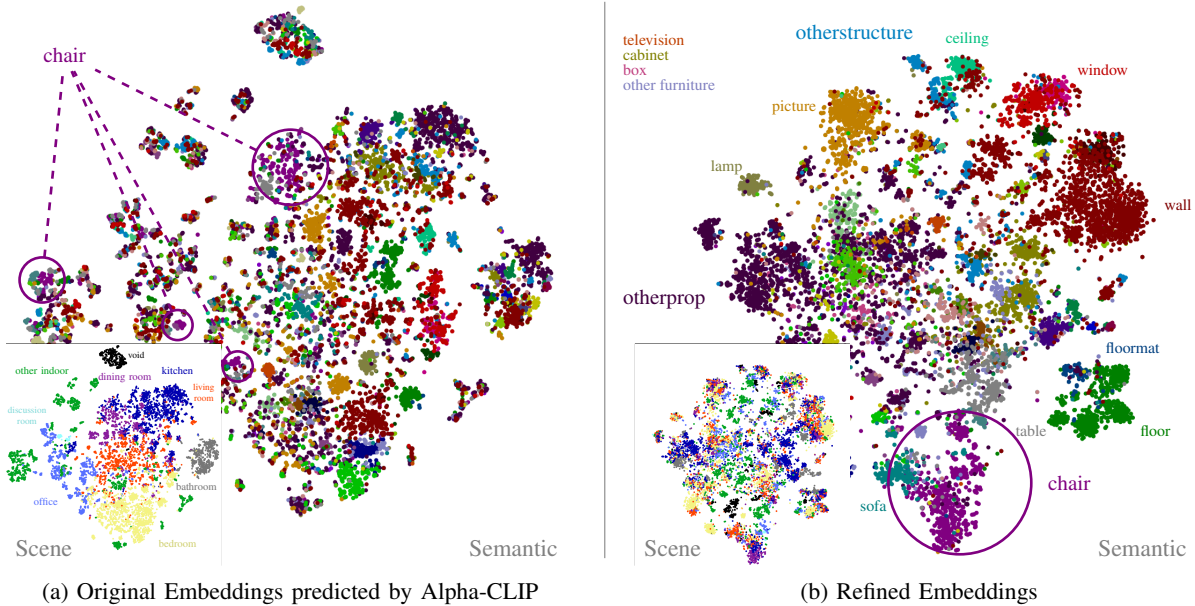
\begin{figure*}
    \vspace{0.8mm}%
    \centering
    \tikzset{tsne_scene/.style={inner sep=0, draw}}
    \tikzset{label/.style={font=\small, gray}}
    \tikzset{figlabel/.style={font=\small}}
    \tikzset{highlight/.style={highlight_color, line width=1px, draw, inner sep=0, outer sep=0, circle}}
    \begin{tikzpicture}
        \phantomsubcaption\label{fig:t-sne::sem}%
        \phantomsubcaption\label{fig:t-sne::sem_diff}%
        \node (tsne_sem) at (0, 0) {
            \includegraphics[width=0.4\hsize]{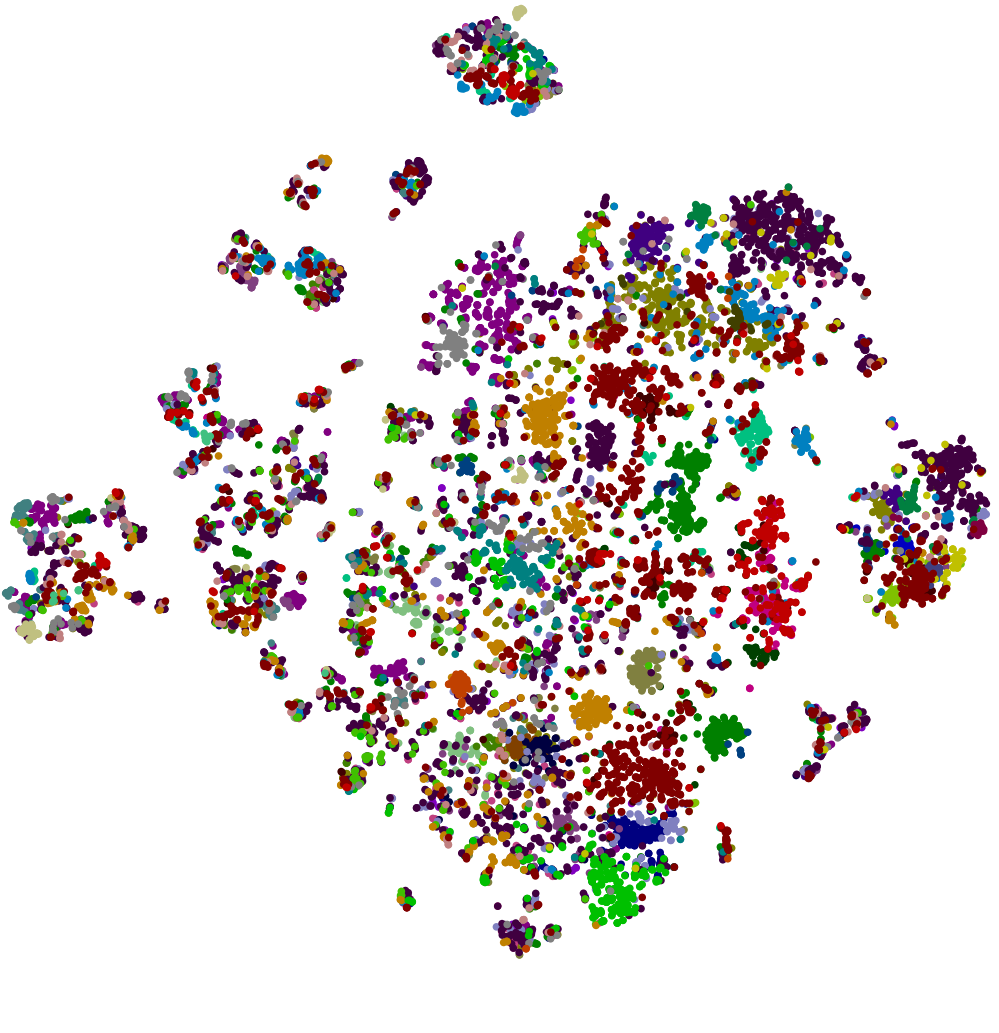}
        };
         \node[tsne_scene, anchor=south west] (tsne_scene) at ($(tsne_sem.south west)-(0.1, 0)$){
            \includegraphics[width=0.16\hsize]{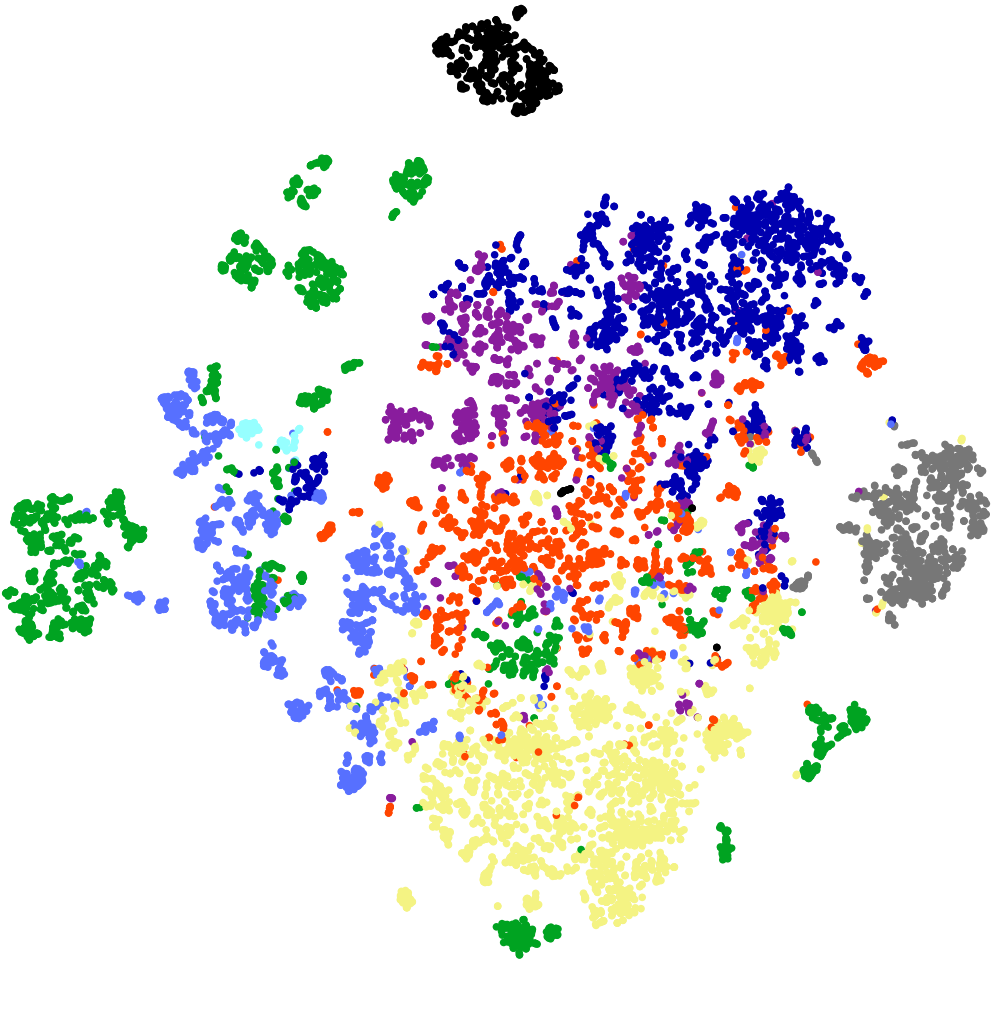}
        };

        \node[anchor=south west, label] at (tsne_scene.south west) {Scene};
        \node[anchor=south east, label] at (tsne_sem.south east) {Semantic};

        \node[right= of tsne_sem, anchor=west] (tsne_sem_diff){
            \includegraphics[width=0.4\hsize]{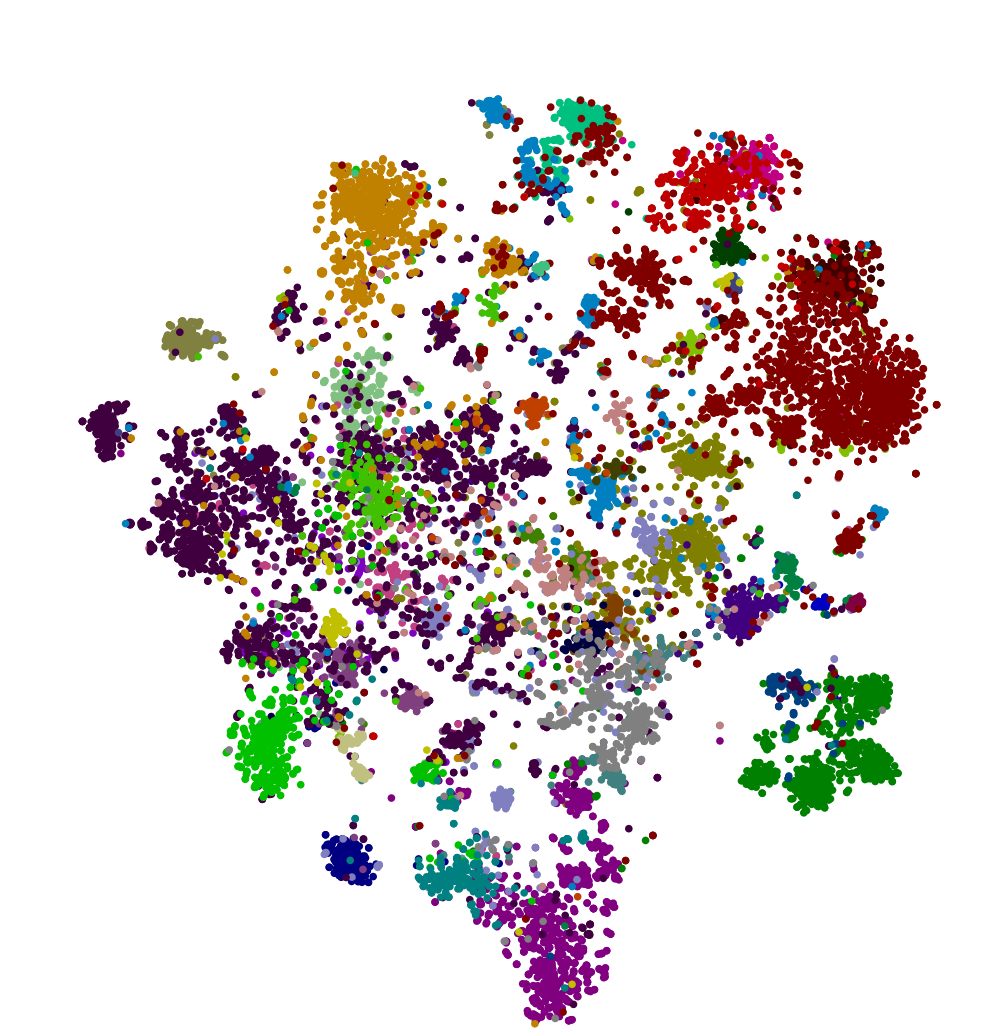}
        };
        \node[tsne_scene, anchor=south west] (tsne_scene_diff) at ($(tsne_sem_diff.south west)-(0.1, 0)$) {
            \includegraphics[width=0.16\hsize]{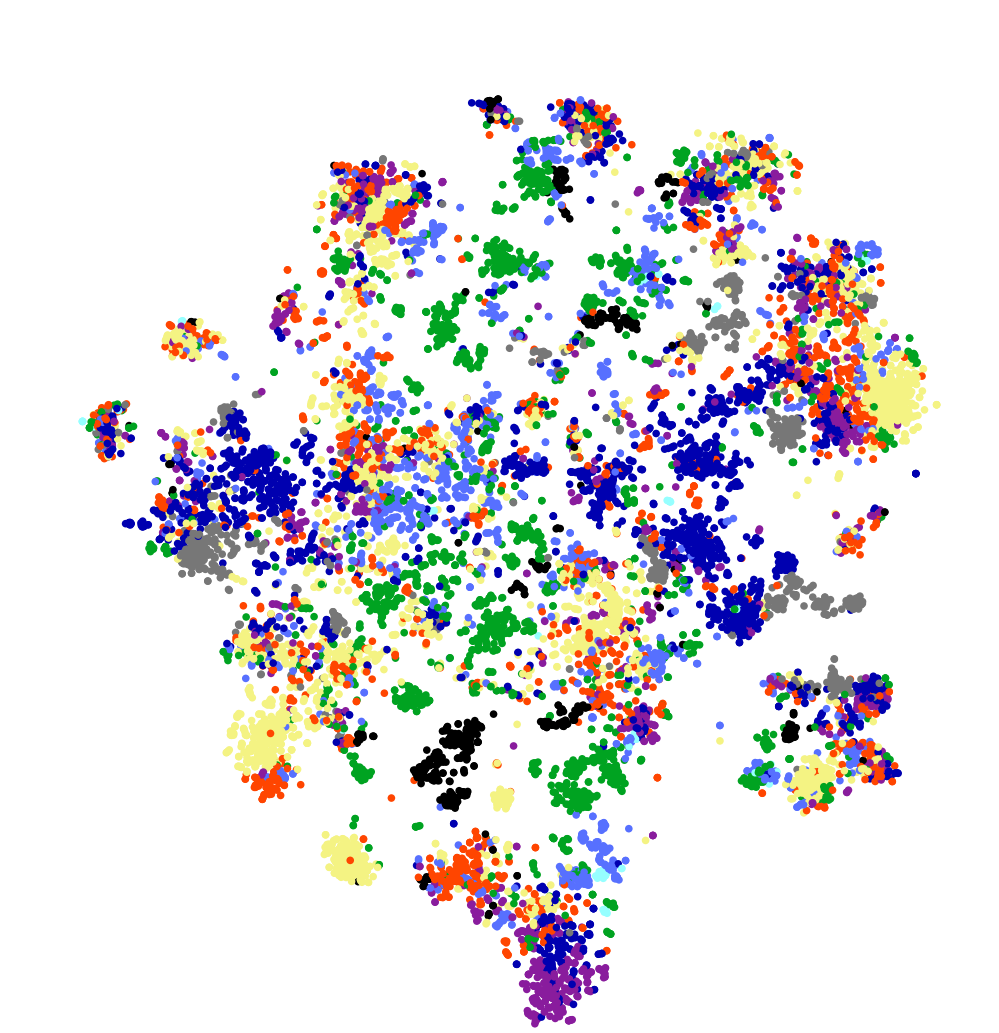}
        };

        \node[anchor=south west, label] at (tsne_scene_diff.south west) {Scene};
        \node[anchor=south east, label] at (tsne_sem_diff.south east) {Semantic};

        \draw[gray] let 
            \p1=($(tsne_sem)!0.5!(tsne_sem_diff)$),
            \p2=($(tsne_sem.north)-(0, 0.1)$)
        in
            (\x1, -\y2) -- (\x1, \y2);

        \node[below=0.1 of tsne_sem, figlabel] {%
            \subref*{fig:t-sne::sem}~Original Embeddings predicted by Alpha-CLIP%
        };%
        \node[below=0.1 of tsne_sem_diff, figlabel] {%
            \subref*{fig:t-sne::sem_diff}~Refined Embeddings%
        };%

        \node[highlight, minimum size=1cm] (ch1) at (0, 1.5) {};
        \node[highlight, minimum size=0.6cm] (ch2) at (-3.3, 0) {};
        \node[highlight, minimum size=0.4cm] (ch3) at (-1.5, -0.6) {};
        \node[highlight, minimum size=0.4cm] (ch5) at (-0.7, -1.15) {};

        \node[highlight, minimum size=1.9cm] (ch4) at (8.9, -2.7) {};

        \node[highlight_color, font=\footnotesize] (ch_cluster) at (-3, 3) {chair};
        \node[highlight_color, font=\footnotesize, right= 0.1 of ch4] (ch_cluster2) {chair};

        \draw[highlight_color, dashed, line width=1px] (ch1) -- (ch_cluster);
        \draw[highlight_color, dashed, line width=1px] (ch2) -- (ch_cluster);
        \draw[highlight_color, dashed, line width=1px] (ch3) -- (ch_cluster);
        \draw[highlight_color, dashed, line width=1px] (ch5) -- (ch_cluster);

        \node[kitchen_color, font=\tiny, scale=0.8] at (-1.5, -1.2) {kitchen};
        \node[office_color, font=\tiny, scale=0.8] at (-3.25, -2.95) {office};
        \node[bathroom_color, font=\tiny, scale=0.8] at (-1.25, -2.75) {bathroom};
        \node[otherindoor_color, font=\tiny, scale=0.8] at (-3.3, -1.2) {other indoor};

        \node[bedroom_color!85!black, font=\tiny, scale=0.8] at (-1.5, -3.5) {bedroom};
        \node[diningroom_color, font=\tiny, scale=0.8] at (-2.2, -1.3) {dining room};
        \node[livingroom_color, font=\tiny, align=center, scale=0.7] at (-1.15, -1.5) {living\\room};
        \node[discussionroom_color!85!black, font=\tiny, align=center, scale=0.7] at (-3.5, -1.8) {discussion\\room};
        \node[black, font=\tiny, align=center, scale=0.7] at (-2.05, -1) {void};

        \node[otherprop_color, font=\footnotesize, anchor=north west] at (4.3, 0) {otherprop};
        \node[otherstructure_color, font=\footnotesize] at (7.5, 3.3) {otherstructure};
        \node[window_color, font=\ssmall] at (10.8, 2.8) {window};
        \node[wall_color, font=\ssmall] at (11.8, 1) {wall};
        \node[floor_color, font=\ssmall] at (11.5, -2) {floor};
        \node[picture_color, font=\ssmall] at (6.7, 2.3) {picture};
        \node[ceiling_color, font=\ssmall] at (9, 3.2) {ceiling};
        \node[sofa_color, font=\ssmall] at (7.75, -3) {sofa};
        \node[table_color, font=\ssmall] at (9.85, -2) {table};
        \node[lamp_color, font=\ssmall] at (5.8, 1.5) {lamp};
        \node[floormat_color, font=\ssmall] at (11.2, -0.9) {floormat};

        \node[tv_color, font=\ssmall, anchor=north west] (tv) at (4.3, 3.4) {television};
        \node[cabinet_color, font=\ssmall, anchor=north west] at  (4.3, 3.2) {cabinet};
        \node[box_color, font=\ssmall, anchor=north west] at  (4.3, 3) {box};
        \node[otherfurniture_color, font=\ssmall, anchor=north west] at  (4.3, 2.8) {other furniture};
        
    \end{tikzpicture}%
    \vspace{-1.5mm}%
    \caption{%
        Visualization of segment embeddings generated by Alpha‑CLIP (see Sec.~\ref{sec:embedding_vector_estimation}) for the NYUv2 training split: %
        \subref*{fig:t-sne::sem}~shows the embeddings before removing global scene context, while \subref*{fig:t-sne::sem_diff}~depicts them after context suppression~($\alpha=0.65$). %
        In both figures, in the main view, points are colored by the ground-truth semantic class, and the subplot shows the same points colored by their scene class~(as defined in~\cite{emsanet2022ijcnn}). %
        Note that in \subref*{fig:t-sne::sem}~clusters form primarily by scene class, whereas in \subref*{fig:t-sne::sem_diff}, due to scene context suppression, clusters are more distinctly organized by semantic class.
        The embeddings were reduced using PCA and t‑SNE using cosine similarity as the distance metric.%
    }%
    \label{fig:t-sne}%
    \vspace{-3.5mm}%
\end{figure*}%
This behavior\footnote{Also discussed in Alpha-CLIP GitHub Issue: {\color{gray}\url{https://github.com/SunzeY/AlphaCLIP/issues/67}}.} can be understood as an application-dependent trade-off between local object vs. global scene focus.
Our application, however, is more dependent on distinguishing each segment's specific attributes (e.g., object class, color or material properties) rather than on global scene cues.
To address this, we refine $\seMi$ by suppressing scene context.
To do so, we compute a global image embedding $\seI$ using Alpha-CLIP by treating the whole image as a single large segment, and adjust each mask embedding as:
\begin{equation}%
    \steMi = \frac{\seMi}{\|\seMi\|} - \alpha \cdot \frac{\seI}{\|\seI\|}
\end{equation}%
The parameter~$\alpha$ controls how much global scene context is removed, allowing the segment embeddings to become more segment-centric while still retaining some scene information if desired.
As shown in Fig.~\ref{fig:t-sne::sem_diff}, this adjustment forms clusters of embeddings with semantic categories, rather than scene classes.
Once computed, these adjusted segment embeddings $\steMi$ are stored and later used for training. 

\subsection{Model Architecture and Knowledge Distillation}%
\label{sec:model_and_knowledge_distillation}%
For DVEFormer, we build upon EMSAFormer~\cite{emsaformer2023ijcnn}~--~an RGB-D encoder-decoder model originally designed for multiple mobile-robotics tasks, such as panoptic segmentation~(by combining semantic and instance segmentation), instance orientation estimation, and scene classification.
In this paper, we exclusively focus on the semantic segmentation branch and build upon its architecture to predict dense pixel-wise visual embeddings. 
EMSAFormer uses a modified Swin-V2-T backbone~\cite{swinv2-cvpr2022}, which has been extended in~\cite{emsaformer2023ijcnn} to jointly process RGB and depth inputs in a single encoder.
For semantic segmentation, EMSAFormer explored two alternative decoders~--~a lightweight SegFormer-like~\cite{SegFromer-neurips2021} MLP-based decoder and the decoder of ESANet~\cite{esanet2021icra}.
We build upon the MLP-based decoder due to its slightly better efficiency and strong performance on large-scale indoor datasets.
Unlike the standard EMSAFormer output, which produces segmentation maps with $40$~channels~(for the $40$~semantic NYUv2 classes), our objective is to generate a $768$-dimensional embedding for each pixel.
To account for this higher-dimensional output, we increase the decoder bottleneck from 128 to 512 channels, ensuring sufficient capacity to capture fine-grained visual details.
For more details on the architecture, we refer to our GitHub repository.

To train our DVEFormer to predict dense pixel-wise visual embeddings, we adopt a knowledge distillation strategy.
For each dataset sample, teacher embeddings~$\{\steMi\}$ are extracted offline using Alpha-CLIP~(see Sec.~\ref{sec:embedding_vector_estimation}).
Subsequently, the embeddings are combined according to their corresponding segment masks~$\{\sMi\}$~(derived from panoptic annotations) to form a teacher volume~$\sY\in\mathbb{R}^{H\times{}W\times{}D}$. 
As the segment masks are unique and non-overlapping, there is no ambiguity in the teacher target for each pixel.
Since CLIP-based models typically align visual and textual features via cosine similarity, we decided to use cosine distance as loss.
Specifically, the student’s predicted embeddings~$\sYh\in\mathbb{R}^{H\times{}W\times{}D}$ are compared to its corresponding teacher embeddings $\sY$ in a pixel-wise manner:
\begin{equation}%
    \mathcal{L} = \frac{1}{|\mathcal{P}|}\sum_{(x,y)\in\mathcal{P}} \Bigl(1 - \frac{\sYxy\cdot\sYhxy}{||\sYxy||\cdot||\sYhxy||}\Bigr)%
\end{equation}%
where $\mathcal{P}$ denotes the set of pixels covered by any binary segment mask, and $\sYxy\in\mathbb{R}^D$ is the embedding in $\sY$ at pixel position $(x, y)$.
Aligning the student's dense output with these teacher embeddings in this way helps to preserve alignment to the output of the text encoder, enabling downstream text-guided tasks~-~though this depends on training with a sufficiently large dataset. 

While the teacher model processes a single RGB image and one alpha mask at a time to generate a segment embedding, requiring repeated inference for multiple segments, our student model generates dense pixel-wise embeddings from RGB and depth inputs and, thus, must implicitly learn to identify the segments without access to the masks.
This difference creates a domain gap, as the student needs to both segment and embed simultaneously.
Using the additional depth input can provide geometric cues -- such as object boundaries -- that can help to bridge this gap and refine the student's predictions.

\subsection{Dense Visual Embeddings in Application}%
\label{sec:model_interpret_embedding}%
Our model predicts dense embeddings, encoding visual information that can be used in various downstream applications.
For example, as our training objective aligns the predicted embeddings with Alpha-CLIP’s joint vision–text space, one can leverage the provided text encoder to embed a prompt~(or several prompts) that describe an object of interest.
Each pixel’s embedding can then be compared via cosine similarity to the text embedding, so that pixels with higher similarity indicate a closer match to the prompt~(see Fig.~\ref{fig:approach::application_querry}).  %
Further thresholding the similarities results in a fine-grained segmentation mask.
The predicted embeddings can also be converted into closed-set predictions without any additional training to accomplish classical semantic segmentation.
One approach is to generate representative embeddings for each predefined semantic class.
This can be done by using a prompt-based strategy (e.g., \textit{``A picture of a \{semantic\_class\_name\}``}) to obtain a class embedding~(see Fig.~\ref{fig:approach::application_text_based}) or by computing the mean embedding for all segments belonging to the same semantic class on the training set~(referred to as visual mean, see Fig.~\ref{fig:approach::application_visual_mean}).
In both cases, a pixel-wise assignment is made by selecting the class with the highest cosine similarity~(arg$\,$max). %
For applications that require an even stronger performance on a specific dataset, a linear probing strategy~\cite{clip-icml2021} can be used~(see Fig.~\ref{fig:approach::application_linear_probing}).
Here, an additional linear layer is trained~--~while keeping the rest of the model frozen~--~to map the $768$-dimensional embedding of a single pixel directly to the set of predefined classes and, thereby, adapting the representation to the target dataset.
Although this requires some additional training, it does not necessitate retraining the entire model, making it a fast and lightweight adaptation method.

Beyond single image-based classification tasks, dense embeddings can be directly utilized for other robotic applications, such as 3D mapping and navigation, where preserving a continuous representation of the scene is advantageous over discretization to selected predefined classes.

These described applications highlight the flexibility of our approach within an application stack.
Some applications may favor closed-set predictions for consistency or integration with existing systems, while others might benefit from retrieval using aligned text embeddings or visual reference embeddings (e.g. visual mean).
With DVEFormer, all these functionalities can be achieved seamlessly from the same dense embedding output, with little overhead. 
\section{Experiments}
\label{sec:experiments}
Our experimental evaluation is conducted on several widely used indoor segmentation datasets, including NYUv2~\cite{NYUv2-eccv2012}, SUN RGB-D~\cite{SUNRGBD-cvpr2015}, and ScanNet~\cite{scannet-cvpr2017}.
We begin by focusing on the smaller NYUv2 dataset and aligning our model's output solely to the teacher embeddings extracted for this dataset to validate the feasibility of our approach.
We conduct several closed-set experiments for classical semantic segmentation under various settings to identify configurations that perform well.
Building upon these results, as we expect our model to perform better with more data, we scale our training to a combined set of segmentation datasets following the training recipe from \cite{panopticndt2023iros} that mixes NYUv2, SUN RGB-D, ScanNet, and Hypersim~\cite{hypersim-iccv2021}, which all share a similar class spectrum.
Furthermore, we make use of the ADE20k dataset~\cite{ade20k-ijcv2019} to further increase the amount of training data\footnote{
    Since ADE20k lacks depth images, we estimated them using Depth-Anything-V2~\cite{depthanythingv2-nips2025}. 
    We used the ViT-L-based model trained on synthetic images from Hypersim for indoor depth estimation. 
}.
Even though it does not share the same class spectrum and is not specifically designed for indoor environments, our approach is not bound to fixed semantic classes, and, thus, the additional data can be easily integrated into our training set, which is an advantage over closed-set approaches.
We evaluate the performance across multiple indoor datasets and compare our approach to other state-of-the-art approaches.
Finally, we present qualitative results that highlight the strengths of our approach.

\subsection{Implementation Details}%
Our implementation builds upon the EMSAFormer~\cite{emsaformer2023ijcnn} codebase.
We followed the same training and data augmentation pipeline as EMSAFormer. 
We initialized our encoder with ImageNet pre-trained weights and used an input resolution of 640${\times}$480.
Training was performed on an NVIDIA A100~(40~GB), however, due to the high memory demands of certain configurations, we reduced the batch size to 4.
In contrast to EMSAFormer, we utilized automatic mixed precision (AMP) with bfloat16 to accelerate training and reduce memory consumption
We used AdamW~\cite{adamw-2019iclr} as optimizer and conducted a grid search over learning rates $\{4\mathrm{e}{-5}, ~3\mathrm{e}{-5}, ~2\mathrm{e}{-5}, ~1\mathrm{e}{-5},~9\mathrm{e}{-6}\}$ with an additional one-cycle learning rate scheduler.
Depending on the dataset setting, the model was trained for either 500 or 250 epochs.
For linear-probing experiments, the entire model, except for an additional linear layer at the output that projects the 768-dimensional output embeddings to a fixed set of semantic classes, was frozen.
This layer was trained for \nicefrac{1}{10} of the total epochs of the initial training at a learning rate of $1\mathrm{e}{-3}$, while keeping the rest of the training protocol unchanged.
For generating teacher embeddings, we used the Alpha-CLIP~\cite{alphaclip2024cvpr} model based on the original CLIP-L/14@336 checkpoint~\cite{clip-icml2021} and fine-tuned on GRIT-20M~\cite{grit-arxiv2023} by the Alpha-CLIP team.
For experiments involving a text encoder, we employed the one associated with the used Alpha-CLIP checkpoint.
More details regarding the data pipeline and training parameters are available in our GitHub repository.

\subsection{Evaluation and Metrics}%
To the best of our knowledge, ground-truth data for indoor applications beyond the datasets' closed-set annotations are not available for evaluating all aspects of our model.
Therefore, we evaluate the performance by reporting the mean intersection over union~(mIoU) for the three complementary scenarios detailed in Sec.~\ref{sec:model_interpret_embedding}. 

\textit{Text-based:} \enspace %
Predicted embeddings are assigned with the text embeddings of the corresponding classes (e.g., \textit{``A picture of a shelf''}) based on the cosine similarity.
This approach is especially useful when the segmentation classes are not known beforehand or may depend on user input.

\textit{Visual mean:} \enspace %
The pixel assignment is done by comparing predicted embeddings to the mean embedding of each class derived from the training set. 
This setting reflects a scenario where an object of interest is observed multiple times. 
Its mean embedding can be stored and serve as a reference embedding, all without requiring additional training.

\textit{Linear probing:} \enspace %
A linear layer is trained to map the dense outputs to the fixed spectrum of semantic classes, yielding a robust measure for closed-set performance.

\subsection{Results on NYUv2}
\label{sec:nyuv2_results}
\begin{figure}[!t]%
    \vspace{1mm}%
    \centering
    \includegraphics[width=\linewidth]{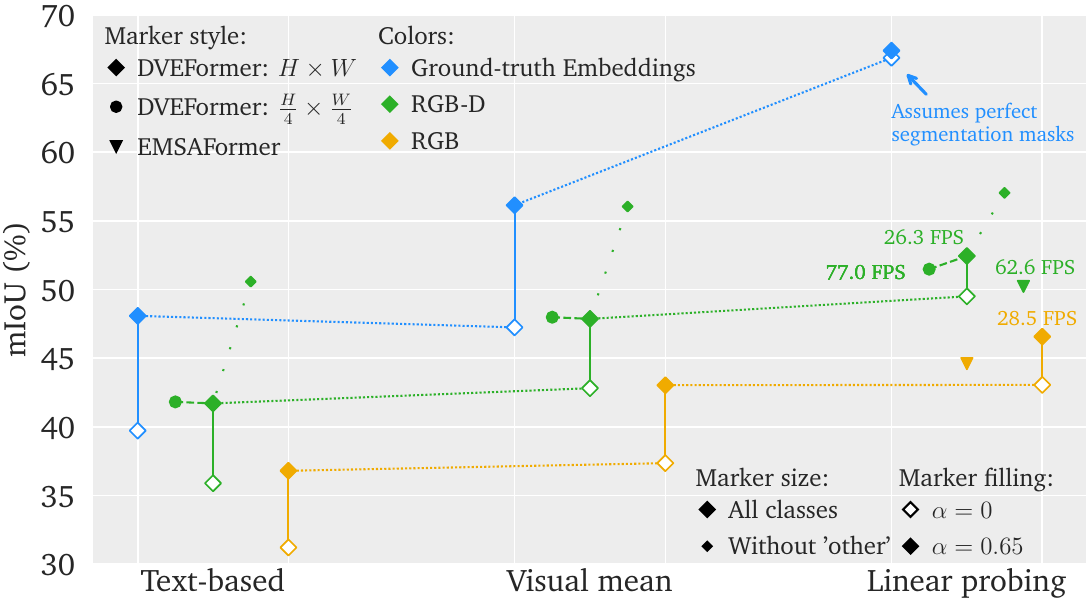}%
    \vspace{-1mm}%
    \caption{%
        Closed-set results for classical semantic segmentation on the NYUv2 test split for different configurations~(see all legends and the x-axis). 
        Note that ground-truth masks are used for computing the GT performance and, thus, assume perfect segmentation.
        Therefore, the gap between the learned models and the GT is not solely due to the knowledge distillation process.
        Throughput was measured on an NVIDIA Jetson AGX Orin~(JetPack 6.2, TensorRT 10.3, Float16). 
        Additional details on the metrics and experimental setup are provided in Sec.~\ref{sec:nyuv2_results}.
    }%
    \label{fig:nyuv_experiments}%
    \vspace{-4mm}%
\end{figure}

We begin by evaluating our proposed approach on NYUv2, elaborating key aspects of our approach. 
The results are summarized by Fig.~\ref{fig:nyuv_experiments}.
The ground-truth~(GT) performance in this figure refers to assigning the target embeddings~${\sY}$ as predictions and, subsequently, computing the mIoU.
We report this for completeness, serving as a theoretical upper bound.
However, it should be noted that the GT performance assumes perfect segmentation masks being available beforehand, which is a different scope compared to evaluating predicted dense embeddings.
Our results indicate that using $\alpha=0.65$ to remove global scene context and focus on individual segments consistently improves mIoU compared to $\alpha=0$.
It can be further seen that RGB-D inputs lead to increased performance compared to RGB inputs, verifying the benefit of using depth data as an additional input for indoor segmentation.
When comparing the closed-set evaluation strategies, linear probing achieves the highest mIoU, even surpassing EMSAFormer’s performance.
Using visual mean embeddings computed from training data consistently outperforms text prompts, indicating that text descriptions may be less optimal to capture the desired visual embeddings, even though they still give suitable results.
Moreover, excluding the three ambiguous classes~(\textit{otherstructure}, \textit{otherfurniture}, and \textit{otherprop}) from NYUv2, as done for the SUN RGB-D dataset~\cite{SUNRGBD-cvpr2015}, improves mIoU for both text-based and visual mean approaches (see Fig.~\ref{fig:nyuv_experiments}).
This coincides with Fig.~\ref{fig:t-sne}, where these classes fail to form distinct clusters, indicating that they are not well-defined and difficult to separate.

Although predicting $768$-dimensional pixel-wise embeddings leads to higher computational cost than EMSAFormer’s $40$-channel output, our model still meets our real-time requirements ($\geq 20$ FPS) on an NVIDIA Jetson AGX Orin.
Moreover, for applications requiring higher throughput, our experiments show that reducing the upsampling factor to $\nicefrac{1}{4}$ of the input resolution significantly boosts FPS (i.e., from $26.3\,$FPS to $77.0$~FPS) with only a slight decrease in accuracy, still outperforming EMSAFormer.
\subsection{State-of-the-Art Comparison}%
Tab.~\ref{tab:dataset_comparison} shows the performance of our model trained on the combined datasets (see introduction to Sec.~\ref{sec:experiments}) when evaluated on the most commonly used real-world indoor segmentation datasets.
Although our main goal is to predict dense text-aligned visual embeddings rather than maximizing closed-set performance, our linear probing approach achieves competitive results, consistently outperforming EMSAFormer and matching the performance of other similarly sized models. 
Notably, while competing approaches~--~except for the last EMSANet row in Tab.~\ref{tab:dataset_comparison}~--~are entirely end-to-end fine-tuned on the target dataset, we only train the additional linear-probing layer on top of our frozen model.
This not only simplifies the training process but also demonstrates that our distilled embeddings can support segmentation with minimal further optimization.
In addition, while text-based and visual mean mIoU values~(gray in Tab.~\ref{tab:dataset_comparison}) are slightly lower than those achieved via linear probing, they still demonstrate decent performance given the flexibility they offer.
Furthermore, our experiments show that our model with reduced upsampling at the output and, thus, faster inference consistently delivers strong performance across all datasets with only a small decrease compared to the full-resolution version.
Overall, these results demonstrate that our approach achieves competitive segmentation performance across diverse indoor datasets while offering flexibility and efficiency suitable for real-world applications.

\begin{table*}[!t]
\vspace{1.2mm}%
\centering
\caption{%
    State-of-the-art comparison for closed-set semantic segmentation on various datasets. %
    For DVEFormer, we report results using linear probing in black and results using text-based\,/\,visual mean in gray.
    For NYUv2 and SUN RGB-D, we report metrics on the test split.
    For ScanNet, validation and hidden test results are included.
    Throughput was measured on an NVIDIA Jetson AGX Orin~(JetPack 6.2, TensorRT 10.3, Float16) with input resolution of 640${\times}$480.
    Legend: $^*$~--~results with test-time augmentation; $\triangledown$~--~expected to be lower; and $^{**}$~--~~deviating input resolution of 1024${\times}$768.
}%
\label{tab:dataset_comparison}%
\vspace{-1.5mm} %
\resizebox{\hsize}{!}{%
    \begin{tabular}{%
        @{}l%
        @{\hspace{4mm}}l%
        @{\hspace{2mm}}%
        c%
        @{\hspace{12mm}}%
        c%
        @{\hspace{4mm}}%
        c%
        @{\hspace{4mm}}%
        c%
        @{\hspace{4mm}}%
        c%
        @{\hspace{4mm}}%
        c%
        @{\hspace{4mm}}%
        c@{}%
    }%
    \toprule
    \textbf{Model}                                                   & \textbf{Backbone}             & \textbf{Extra}              & \textbf{NYUv2}                  & \textbf{SUN RGB-D}              & \textbf{ScanNet}                & \textbf{ScanNet (20 Classes)}   & \textbf{ScanNet (20 Classes)}  & \\
        \vspace{1.5mm}                                                 &                               & \textbf{Training Data}      & \textbf{Test}                   & \textbf{Test}                   & \textbf{Valid}                  & \textbf{Valid}                  & \textbf{Test}          & \\
                                                                     &                               &                             & \textbf{mIoU$^\uparrow$}        & \textbf{mIoU$^\uparrow$}        & \textbf{mIoU$^\uparrow$}        & \textbf{mIoU$^\uparrow$}        & \textbf{mIoU$^\uparrow$}            & \textbf{FPS$^\uparrow$}\\
    \midrule
    ShapeConv~\cite{shapeconv-iccv2021}                              & ResNet-101                    &                             & 47.4\phantom{0}                 & 47.6\phantom{0}                 & -                               & -                               & -                               & \textcolor{gray}{- $\triangledown$} \\
                                                                     & ResNext-101                   &                             & 50.2\phantom{0}                 & -                               & -                               & -                               & -                               & \textcolor{gray}{- $\triangledown$} \\[2mm]
    Omnivore~\cite{omnivore-cvpr2022}                                & Swin-T                        &  \checkmark                 & 49.7\phantom{0}                 & -                               & -                               & -                               & -                               & \textcolor{gray}{- $\triangledown$} \\
                                                                     & Swin-B                        &  \checkmark                 & 54.0\phantom{0}                 & -                               & -                               & -                               & -                               & \textcolor{gray}{- $\triangledown$} \\[2mm]
    MultiMAE~\cite{multi-view-open-vocab-2023prcv, dformer-iclr2024} & ViT-B                         & \checkmark                  & 56.0\phantom{0}                 & 51.1\phantom{0}                 & -                               & -                               & -                               & \textcolor{gray}{- $\triangledown$}\\[2mm]
    CMX~\cite{cmx-its2023}                                           & 2x MiT-B2                     &                             & 54.1\phantom{0}                 & 49.7$^*$                        & -                               & -                               & 61.3                               & 33.37\\      
                                                                     & 2x MiT-B4                     &                             & 56.0\phantom{0}                 & 52.1$^*$                        & -                               & -                               & -                               & 18.76\\[2mm]
    CMNeXt~\cite{cmnext-cvpr2023, dformer-iclr2024}                  & 2x MiT-B4                     &                             & 56.9\phantom{0}                 & 51.9\phantom{0}                 & -                               & -                               & -                               & 14.84 \\[2mm]
    DFormer~\cite{dformer-iclr2024}                                  & DFormer-S                     & \checkmark                  & 53.6\phantom{0}                 & 50.0\phantom{0}                 & -                               & -                               & -                               & 55.03 \\
                                                                     & DFormer-B                     & \checkmark                  & 55.6\phantom{0}                 & 51.2\phantom{0}                 & -                               & -                               & -                               & 35.11 \\[2mm]
    EMSAFormer~\cite{emsaformer2023ijcnn}                            & SwinV2-T-128                  &                             & 50.23                           & 48.61                           & -                               & 64.75                           & 56.4                               & 62.63 \\[2mm]
    EMSANet~\cite{emsanet2022ijcnn}                                  & 2x ResNet34-NBt1D             &                             & 51.15                           & 48.39                           & -                               & -                               & -                               & 78.19\\
    EMSANet~\cite{emsanet2022ijcnn, panopticndt2023iros}             & 2x ResNet34-NBt1D             & \checkmark                  & 56.55                           & 49.31                           & 50.84                           & 66.96                           & 60.00                               & 38.57$^{**}$\\
    \midrule
    \midrule
    \multirow{2}{*}{DVEFormer \textcolor{gray}{($\nicefrac{H}{4}\times\nicefrac{W}{4}$)} (Ours)} & \multirow{2}{*}{SwinV2-T-128} & \multirow{2}{*}{\checkmark} & 56.25                           & 50.11                           & 48.75                           & 67.55            & -        & \multirow{2}{*}{77.0\phantom{0}}\\
           &                               &                             & \textcolor{gray}{43.45 / 50.02} & \textcolor{gray}{44.01 / 45.66} & \textcolor{gray}{33.58 / 39.59} &  \textcolor{gray}{50.32 / 56.75} & \textcolor{gray}{-} &  \\
    \midrule
    \multirow{2}{*}{DVEFormer (Ours)}                                & \multirow{2}{*}{SwinV2-T-128} & \multirow{2}{*}{\checkmark} & 57.07                           & 50.99                           & 49.11                           & 67.72                      & 62.6      & \multirow{2}{*}{26.30} \\
                                                                     &                               &                             & \textcolor{gray}{44.07 / 50.31} & \textcolor{gray}{44.56 / 46.25} & \textcolor{gray}{33.77 / 39.59} & \textcolor{gray}{50.16 / 56.57} & \textcolor{gray}{-} &\\
    \bottomrule%
    \end{tabular}%
}%
\vspace{-6mm}%
\end{table*}
\section{Application of DVEFormer for 3D Mapping}
\label{sec:application}
To further demonstrate the applicability of our approach, we integrate the predicted dense visual embeddings into the existing 3D normal distribution transform~(NDT) mapping framework~\cite{panopticndt2023iros}, originally designed in~\cite{semanticmapping2022icra} to incorporate semantic segmentation.
Fig.~\ref{fig:application-examples} shows an indoor scene alongside its corresponding 3D-NDT map, where each cell stores accumulated visual embeddings from multiple observations instead of per-class histogram counts. 
The same map is used for two applications.
For classical semantic segmentation, each cell’s embedding is normalized and classified using the linear probing weights from our experiments, assigning the class with the highest confidence. 
Additionally, text queries retrieve objects beyond the fixed class spectrum by computing the cosine similarity between each cell’s embedding and the corresponding text embedding.
Overall, these results showcase that our per-image predicted visual embeddings can be seamlessly integrated into downstream applications such as 3D mapping, thereby increasing overall flexibility.

For more impressions and additional examples in real-world applications, we refer to our GitHub repository.
\section{Conclusion}
\label{sec:conclusion}
In this paper, we have presented an approach for efficient prediction of dense text-aligned visual embeddings utilizing RGB-D Transformers, called DVEFormer.
Instead of performing classical semantic segmentation with fixed predefined classes, we employ a knowledge distillation strategy that uses teacher embeddings from Alpha-CLIP to guide our model to learn pixel-wise visual embeddings.
Our evaluations on NYUv2, SUN RGB-D, and ScanNet demonstrate that our method not only achieves competitive closed-set performance via linear probing but also supports flexible natural-language-based querying and, thus, is well suited for various downstream applications. %
Despite using higher-dimensional outputs, our approach meets real-time inference requirements on mobile platforms with limited computational resources, such as the NVIDIA Jetson AGX Orin by reaching $26.3\,$FPS for our larger model and $77.0\,$FPS for our smaller model.
In future work, we plan to explore methods to reduce the VRAM required for predicting embeddings, such as reducing embedding dimensions or predicting binary segment masks along with a single embedding vector per mask.
Additionally, we plan to train on larger datasets, evaluate instance awareness and mapping performance more comprehensively, and extend our approach toward panoptic segmentation.
\definecolor{query_color}{HTML}{E7E7E7}%
\definecolor{sem_color}{HTML}{E7E7E7}%
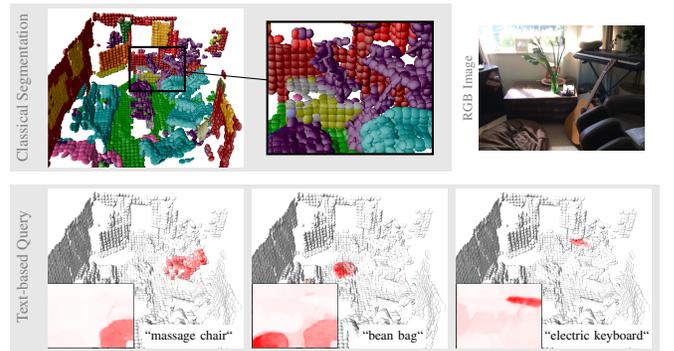
\begin{figure}[!t]%
    \vspace{2mm}%
    \centering%
    \tikzset{image/.style={inner sep=0}}%
    \tikzset{hightlight/.style={line width=1.5px, black, draw}}%
    \tikzset{imgquery/.style={line width=0.5px, draw}}%
    \begin{tikzpicture}[spy using outlines={rectangle, magnification=3, width=2.2cm, height=0.8*2.2cm, connect spies}]%

         \coordinate (query_anchor) at (0, 0);%
         \coordinate (sem_anchor) at (0, 2.4);%
         \def\ndtimgsize{0.3\hsize};%
    
        \node[image, left= 1.41 of sem_anchor] (ndt_sem) {
            \includegraphics[width=0.3\hsize, trim={0, 0, 2px, 0}, clip]{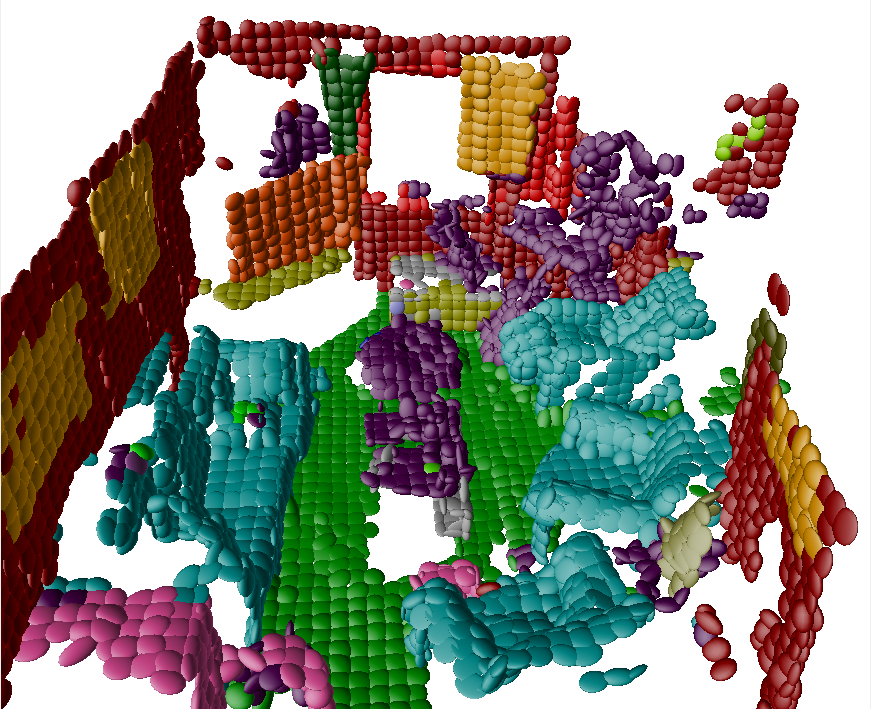}
        };
        \node[left= 0.1 of ndt_sem, font=\ssmall, gray] {\rotatebox{90}{Classical Segmentation}};%

        \spy[black, every spy on node/.append style={thick}] on ($(ndt_sem)+(0.15, 0.25)$) in node at (sem_anchor);%

        \node[image, left=1.41 of query_anchor] (ndt_query_4) {%
            \includegraphics[width=\ndtimgsize]{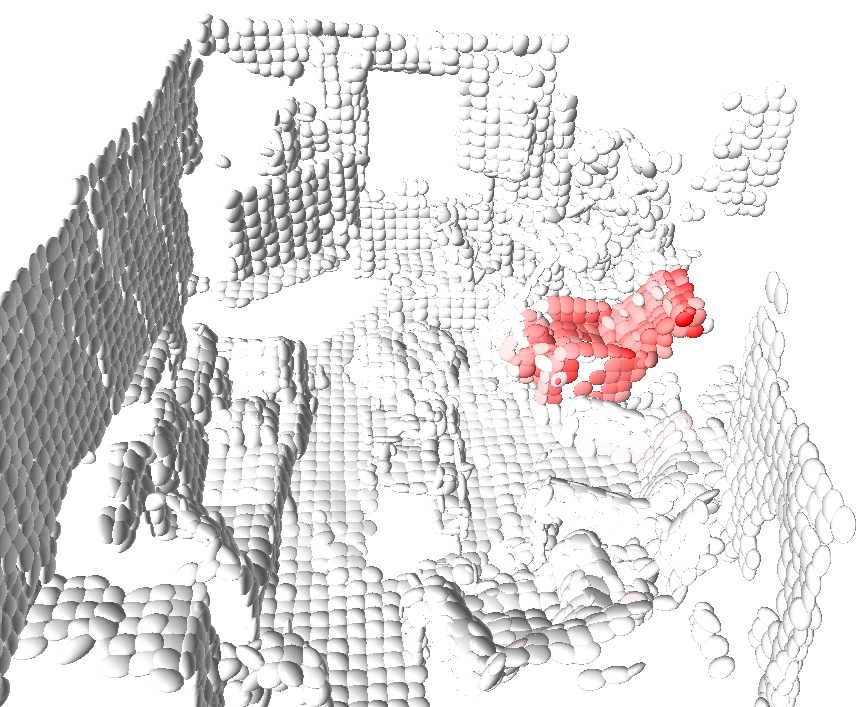}%
        };%
        \node[image, anchor=south west, imgquery] (img_query_4) at (ndt_query_4.south west) {%
            \includegraphics[width=0.13\hsize]{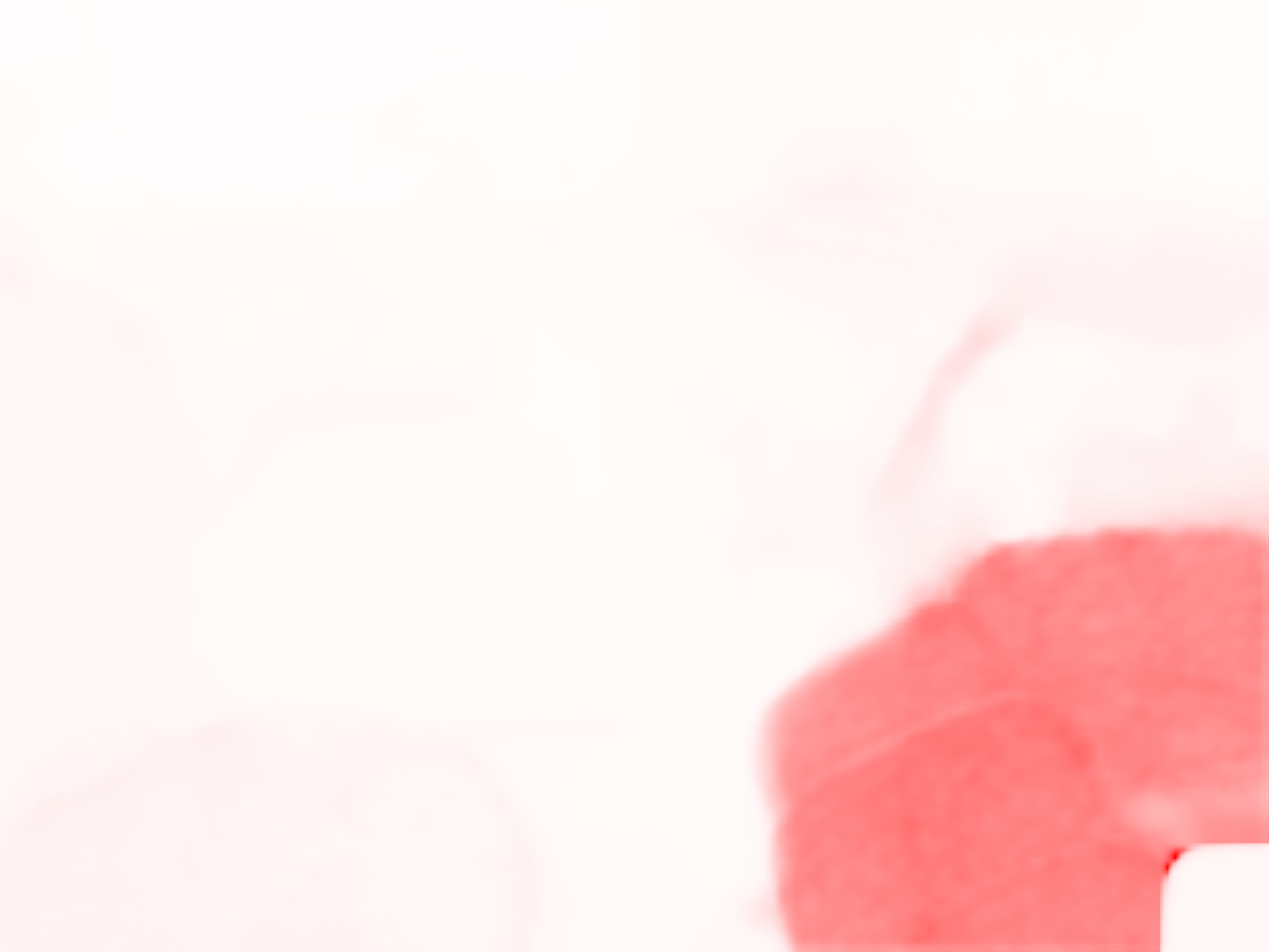}%
        };%
        \draw let%
            \p1=($(ndt_query_4.south east)!0.5!(img_query_4.south east)$)%
        in%
            node[anchor=south, fill=white, font=\tiny, inner sep=1px, minimum height=0.3cm] at (\p1) {%
                ``massage chair``
            };%

        \node[image] at (query_anchor) (ndt_query_2) {%
            \includegraphics[width=\ndtimgsize, trim={0, 0, 2px, 0}, clip]{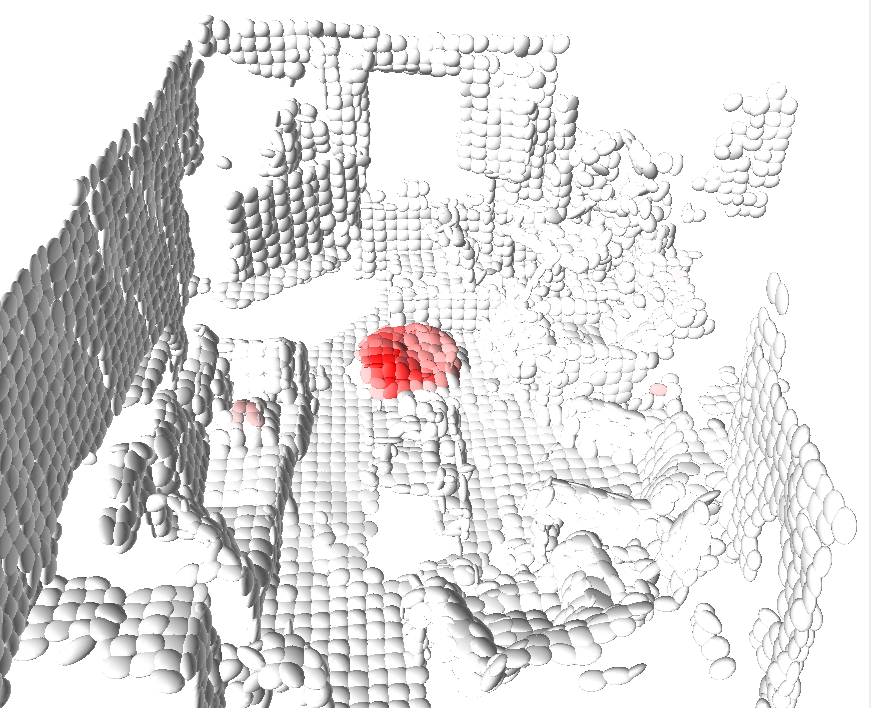}%
        };%
        \node[image, anchor=south west, imgquery] (img_query_2) at (ndt_query_2.south west) {%
            \includegraphics[width=0.13\hsize]{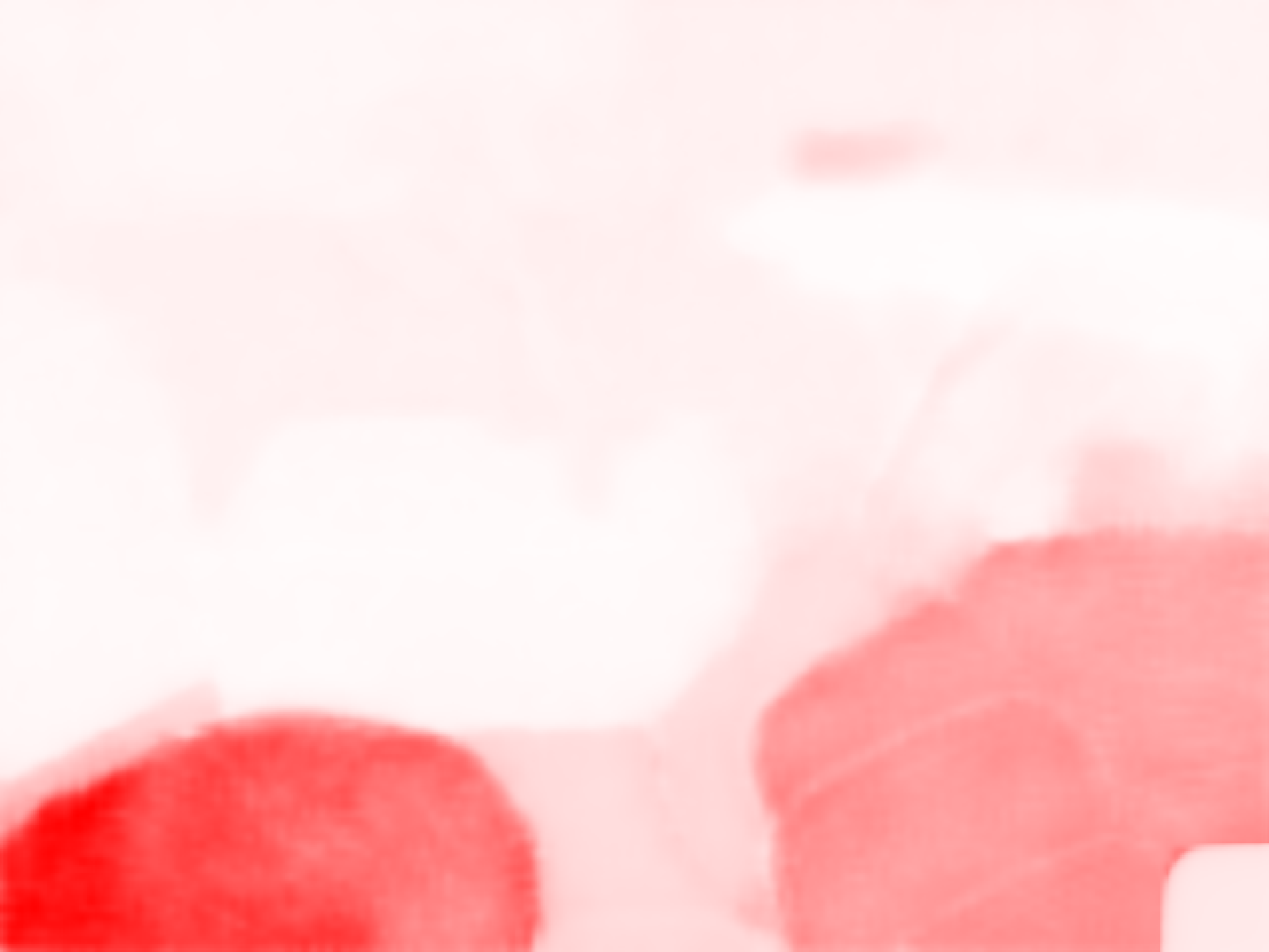}%
        };%
        \draw let%
            \p1=($(ndt_query_2.south east)!0.5!(img_query_2.south east)$)%
        in%
            node[anchor=south, fill=white, font=\tiny, inner sep=1px, minimum height=0.3cm] at (\p1) {
                ``bean bag``
            };%

        \node[image, right=1.41 of query_anchor] (ndt_query_3) {%
            \includegraphics[width=\ndtimgsize, trim={0, 0, 2px, 0}, clip]{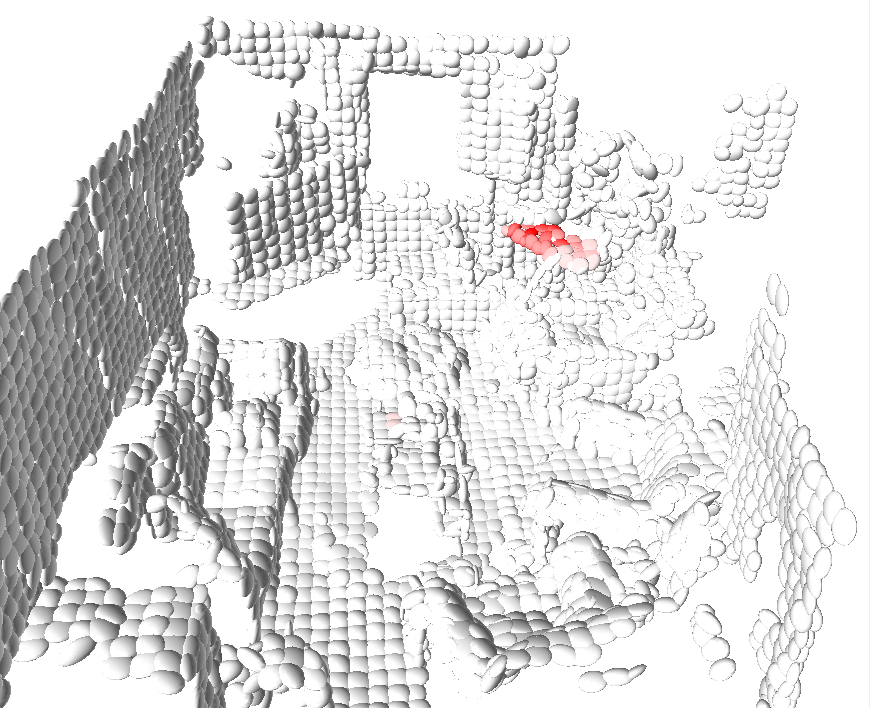}%
        };%
        \node[image, anchor=south west, imgquery] (img_query_3) at (ndt_query_3.south west) {%
            \includegraphics[width=0.13\hsize]{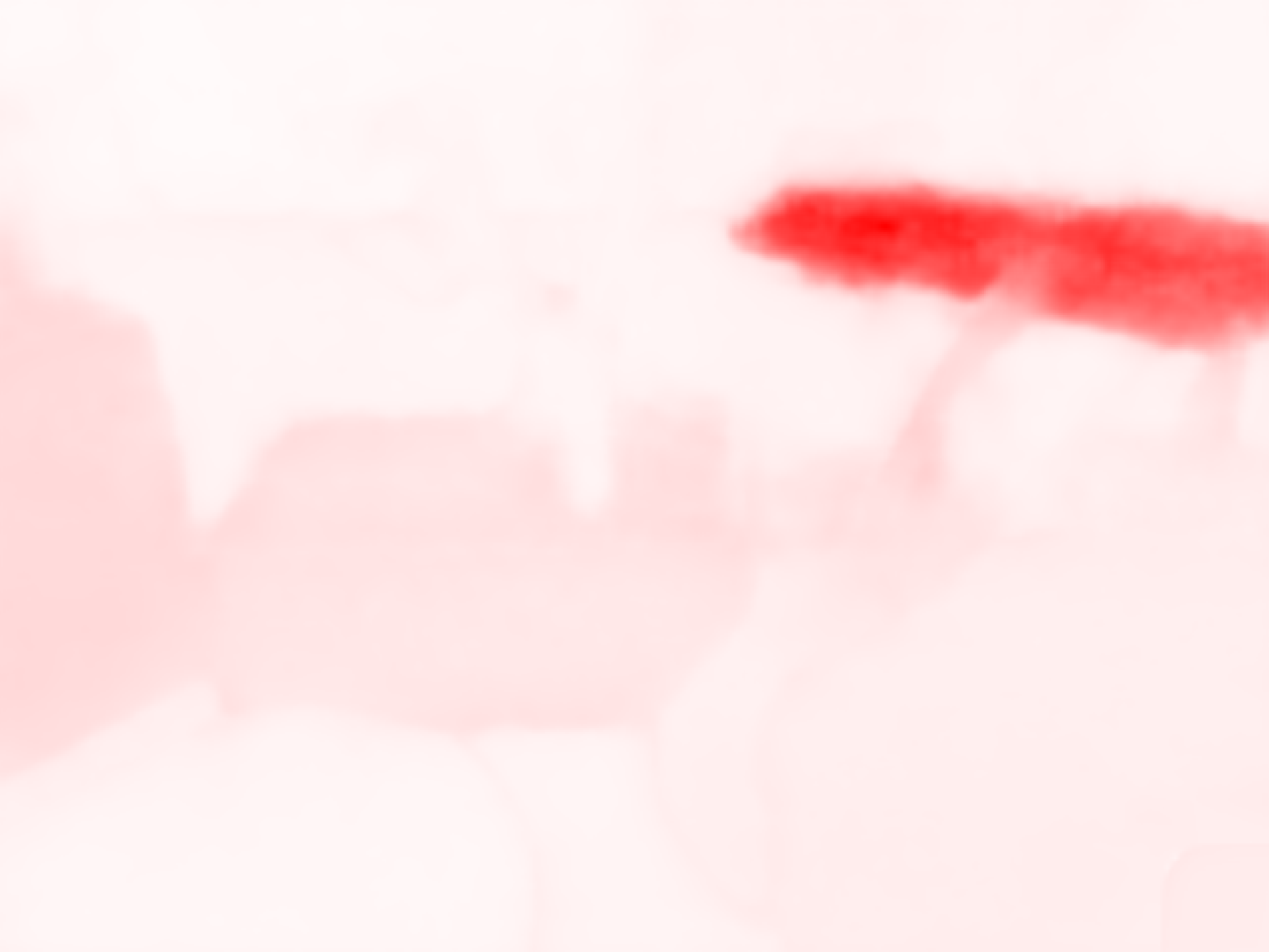}%
        };%
        \draw let%
            \p1=($(ndt_query_3.south east)!0.5!(img_query_3.south east)$)%
        in%
            node[anchor=south, fill=white, font=\tiny, inner sep=1px, minimum height=0.3cm] at (\p1) {``electric keyboard``};%

        \draw let
            \p1=(ndt_sem),
            \p2=($(ndt_query_3)+(0.1, 0)$)
        in
            node[image] (rgb) at (\x2, \y1) {%
                \includegraphics[width=2.2cm]{img/application/00000.jpg}%
            };%

        \begin{scope}[on background layer]%
            \fill[query_color] let%
                \p1=($(ndt_query_4.south west)-(0.5, 0.05)$),%
                \p2=($(ndt_query_3.north east)+(0.1, 0.05)$),%
                \p3=($(ndt_sem)!0.5!(ndt_query_4)$)%
            in%
                (\p1) rectangle (\p2);%
                
            \node[left=0.1 of ndt_query_4, gray, font=\ssmall] {\rotatebox{90}{ Text-based Query}};%
        \end{scope}%

        \begin{scope}[on background layer]%
            \fill[sem_color] let%
                \p1=($(ndt_sem.south west)-(0.5, 0.05)$),%
                \p2=($(ndt_sem.north east)+(0, 0.05)$),%
                \p3=($(ndt_query_2)!0.5!(ndt_query_3)$)%
            in%
                (\p1) rectangle (\x3, \y2);%
        \end{scope}%

        \node[left=0 of rgb, gray, inner sep=1px] {\rotatebox{90}{\tiny RGB Image}};
    \end{tikzpicture}
    \vspace{-5mm}%
    \caption{
        3D-NDT mapping results for a ScanNet validation scene (\textit{valid scene0608}). The map shows classical semantic segmentation (via linear probing) as well as text-based queries (cosine similarity: red: high, white: low). %
        Additionally, an RGB image for a single pose is shown for the same text queries. %
        We refer to Fig.~\ref{fig:t-sne} for the subset of depicted semantic colors. %
    }
    \vspace{-3mm}%
    \label{fig:application-examples}%
\end{figure}%
\bibliographystyle{IEEEtran}
\bibliography{bib/literature.bib}

\end{document}